\documentclass[lettersize,journal]{IEEEtran}

\usepackage{amsmath,amsfonts}
\usepackage{amsthm}
\usepackage{algorithm}
\usepackage{array}
\usepackage[caption=false,font=normalsize,labelfont=sf,textfont=sf]{subfig}
\usepackage{textcomp}
\usepackage{stfloats}
\usepackage{url}
\usepackage{verbatim}
\usepackage{graphicx}
\usepackage{cite}
\usepackage{diagbox} 
\usepackage{amssymb}
\usepackage{algpseudocode}
\usepackage{graphicx}
\usepackage{xcolor}
\usepackage{cases}
\usepackage{multirow}
\usepackage{multicol}
\usepackage{adjustbox}
\usepackage{makecell}
\usepackage{hyperref}
\newtheorem{remark}{Remark}

\begin{document}

\title{KNN-MMD: Cross Domain Wireless Sensing via Local Distribution Alignment}

\author{Zijian Zhao,\IEEEmembership{} Zhijie Cai,\IEEEmembership{} Tingwei Chen,\IEEEmembership{} Xiaoyang Li,\IEEEmembership{} Hang Li\IEEEmembership{}, Qimei Chen\IEEEmembership{}, Guangxu Zhu\IEEEmembership{}
\thanks{Part of the work was presented in 2025 IEEE/CIC International Conference on Communications in China (ICCC) \cite{zhao2025does}.}
\thanks{The work of Guangxu Zhu was supported in part by National Natural Science Foundation of China (Grant No. 62371313),  in part by Shenzhen-Hong Kong-Macau Technology Research Programme (Type C) (Grant No. SGDX20230821091559018), in part by the Shenzhen Science and Technology Program (Grant No. JCYJ20241202124934046), in part by Guangdong Young Talent Research Project (Grant No. 2023TQ07A708). The work of Xiaoyang Li was supported in part by Young Elite Scientists Sponsorship Program by CAST under Grant YESS20240364, Shenzhen Science and Technology Program under Grants JCYJ20241202124934046 and KJZD20240903095402004. The work of Qimei Chen was supported in part by Major Program (JD) of Hubei Province under Grant 2025BEA001, in part by the Wuhan Science and Technology Achievement Transformation Project under Grant 2024030803010178. (Corresponding Author: Guangxu Zhu)}
\thanks{Zijian Zhao is with Shenzhen Research Institute of Big Data, Shenzhen 518115, China, and also with the School of Computer Science and Engineering, Sun Yat-sen University, Guangzhou 510275, China (e-mail: zhaozj28@mail2.sysu.edu.cn)}
\thanks{Zhijie Cai, Tingwei Chen, Hang Li, and Guangxu Zhu are with the Shenzhen Research Institute of Big Data, The Chinese University of Hong Kong (Shenzhen), Shenzhen 518115, China (e-mail:  zhijiecai@link.cuhk.edu.cn; tingweichen@link.cuhk.edu.cn; hangdavidli@163.com; gxzhu@sribd.cn)}
\thanks{Xiaoyang Li is with the Department of Electrical and Electronic Engineering, Southern University of Science and Technology, Shenzhen, 518055, China (e-mail: lixy@sustech.edu.cn)}
\thanks{Qimei Chen is with the School of Electronic Information, Wuhan University, Wuhan, 430072, China (e-mail: chenqimei@whu.edu.cn).}
}


\markboth{IEEE TRANSACTIONS ON MOBILE COMPUTING}%
{Shell \MakeLowercase{\textit{et al.}}: A Sample Article Using IEEEtran.cls for IEEE Journals}


\maketitle

\begin{abstract}
Wireless sensing has recently found widespread applications in diverse environments, including homes, offices, and public spaces. By analyzing patterns in channel state information (CSI), it is possible to infer human actions for tasks such as person identification, gesture recognition, and fall detection. However, CSI is highly sensitive to environmental changes, where even minor alterations can significantly distort the CSI patterns. This sensitivity often leads to performance degradation or outright failure when applying wireless sensing models trained in one environment to another. To address this challenge, {\color{black} Domain Alignment Learning (DAL)} has been widely adopted for cross-domain classification tasks, as it focuses on aligning the global distributions of the source and target domains in feature space. Despite its popularity, DAL often neglects inter-category relationships, which can lead to misalignment between categories across domains, even when global alignment is achieved. To overcome these limitations, we propose K-Nearest Neighbors Maximum Mean Discrepancy (KNN-MMD), a novel few-shot method for cross-domain wireless sensing. Our approach begins by constructing a ``help set" using K-Nearest Neighbors (KNN) from the target domain, enabling local alignment between the source and target domains within each category using Maximum Mean Discrepancy (MMD). Additionally, we address a key instability issue commonly observed in cross-domain methods, where model performance fluctuates sharply between epochs. Further, most existing methods struggle to determine an optimal stopping point during training due to the absence of labeled data from the target domain. Our method resolves this by excluding the support set from the target domain during training and employing it as a validation set to determine the stopping criterion. We evaluate the effectiveness of the proposed method across several cross-domain Wi-Fi sensing tasks, including gesture recognition, person identification, fall detection, and action recognition, using both a public dataset and a self-collected dataset. In a one-shot scenario, our method achieves accuracy rates of 93.26\%, 81.84\%, 77.62\%, and 75.30\% for the respective tasks. 
{\color{black} The dataset and code are publicly available at \href{https://github.com/RS2002/KNN-MMD}{https://github.com/RS2002/KNN-MMD}.}

\end{abstract}


\section{Introduction}
\subsection{Overview}



These days, the importance of wireless sensing has grown significantly, driven by advancements in technology and the increasing demand for efficient monitoring solutions. Wireless sensing has found applications in various fields such as fall detection \cite{fall}, person localization \cite{zhao2024lofi}, and people identification \cite{CSI-BERT}. \textcolor{black}{Among the various manners, Wi-Fi sensing—particularly in the sub-7GHz band—stands out as a popular method with numerous advantages, as outlined in Table \ref{sensing} \cite{zhu2025review}.}\footnote{\color{black}Although Wi-Fi sensing can also be implemented in the millimeter-wave band, this work focuses on sub-7GHz Wi-Fi, which is more widely deployed and better supported by existing open-source CSI extraction tools. Given that millimeter-wave Wi-Fi communication is not yet mainstream \cite{liu2024wi}, all subsequent references to Wi-Fi sensing in this paper refer specifically to sub-7GHz Wi-Fi.} First, Wi-Fi sensing is contactless as the object to be detected does not need to carry a device like a smartphone or smartwatch. \textcolor{black}{Second, Wi-Fi sensing preserves privacy by not capturing visual images or personally identifiable information, in contrast to cameras. Moreover, the raw data it collects is not directly interpretable by humans, further protecting privacy.} Third, Wi-Fi sensing is insensitive to occlusions and illumination conditions\textcolor{black}{, as the sub-7 GHz electromagnetic waves used in Wi-Fi communication can effectively penetrate walls}. Moreover, Wi-Fi sensing is ubiquitous, with over $19.5$ billion Wi-Fi devices deployed globally. With an appropriate algorithm that enables its sensing function, ubiquitous sensing function can be realized in various environments, such as homes, offices, and public spaces. Last but not least, Wi-Fi sensing has lower cost compared to dedicated radars.

\begin{table*}
\caption{\textcolor{black}{Comparison of Common Sensing Methods}}
\begin{adjustbox}{width=\textwidth}
\centering
\begin{tabular}{|c||c|c|c|c|c|}
\hline
& \textbf{Functionality} & \textbf{Accuracy} & \textbf{Coverage} & \textbf{Privacy} & \textbf{Cost} \\
\hline
\textbf{Camera} & \textbf{Highly versatile} & \textbf{High} & Limited (affected by occlusion/lighting) & Low & Low to moderate \\
\hline
\textbf{Millimeter Wave Radar} & Moderately versatile & \textbf{High} & Moderate & \textbf{High} & Expensive \\
\hline
\textbf{Infrared Sensor} & Limited & Medium  & Moderate & \textbf{High} & Low \\
\hline
\textbf{Sub-7 GHz Wi-Fi} & Moderately versatile & Medium  & \textbf{Good (non-line-of-sight, all-day)} & \textbf{High} & \textbf{Very low} \\
\hline
\end{tabular}
\label{sensing}
\end{adjustbox}
\end{table*}

Given the released tool that works with some specific types of commercial Network Interface Cards (NIC) \cite{halperin2011tool}, the Channel State Information (CSI) can be obtained, which describes how the wireless signals fade during the propagation from the sender to the receiver. \textcolor{black}{Using this CSI data collection tool, a series of research employs commercial Wi-Fi NICs for different kinds of tasks.} Moreover, a dedicated IEEE standard for Wi-Fi sensing named IEEE 802.11bf is approved \cite{802.11bf} to enable faster development and practicalization of Wi-Fi sensing technologies. \textcolor{black}{However, since the Orthogonal Frequency Division Multiplexing (OFDM) waveforms employed by Wi-Fi are designed for communication purposes without sensing-specific optimization—such as low-sidelobe ambiguity functions, high time–frequency resolution, or consistent pilot structures—direct range–Doppler modeling from CSI measurements is less reliable than with radar-optimized waveforms, making it challenging to apply physical models to infer the ambient environment.}



\subsection{Cross Domain Challenge}
Recently, most Wi-Fi sensing methods rely on data-driven approach via machine learning, especially for fine-grained tasks like gesture recognition \cite{nirmal2021deep}. However, even minor environmental changes can drastically alter the CSI patterns and significantly degrade the performance of the trained model. This environmental variability leads to a domain shift - a discrepancy between the training data distribution and real-world deployment conditions. Given the strong dependence of the training data, such domain shifts can severely undermine the generalization capability. This scenario is known as a cross-domain problem in machine learning, where the source domain (training data) and target domain (testing data) have different underlying distributions or feature spaces.

In fields like Computer Vision (CV) and Natural Language Processing (NLP), where large datasets are available in these common modalities, it is relatively easy to solve such cross-domain problems by training large models with vast amount of data \cite{chen2024overview}. However, in Wi-Fi sensing, the limited public datasets often have different formats, making it challenging to use them together to train a model. Additionally, collecting large CSI datasets on one's own is not easy. Therefore, it is needed to develop a powerful yet adaptable small model that can quickly adapt to new environments, which has emerged as an important research direction in Wi-Fi sensing.


\color{black}
\subsection{Existing Cross-Domain Wi-Fi Sensing Methods}
\begin{figure}[htbp]
\centering 
\includegraphics[width=0.4\textwidth]{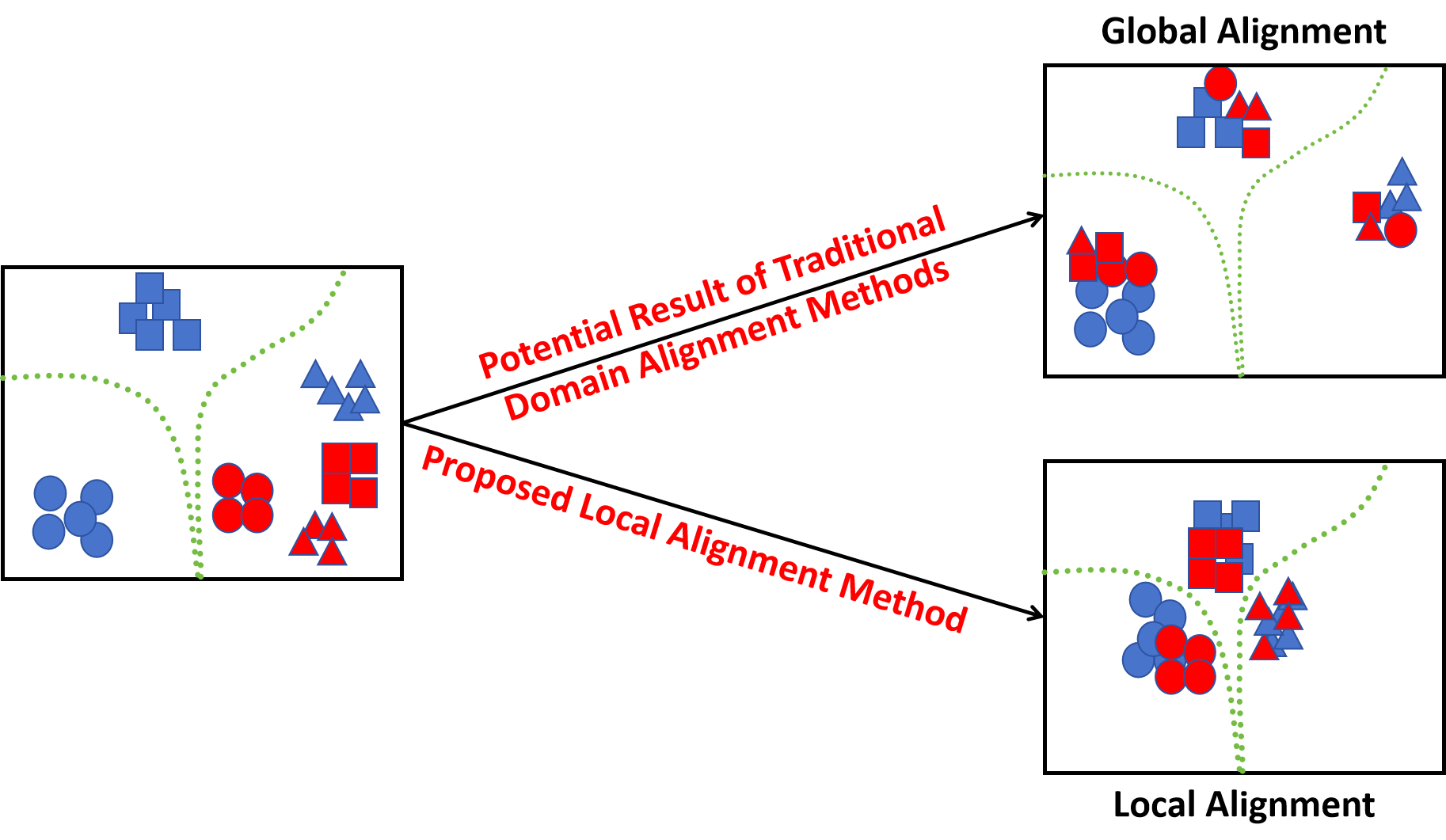}
\caption{Distinguishing Local Alignment from Global Alignment: Different colors represent different domains, different shapes represent different categories, and the green line represents the classification boundary, which remains consistent across subsequent figures.}
\label{fig:performance}
\end{figure}

As the most common method for cross-domain tasks, Domain Adaptation (DA) has been widely utilized in cross-domain Wi-Fi sensing. DA focuses on leveraging knowledge from a source domain to enhance model performance on a related but different target domain. Among these methods, Domain Alignment Learning (DAL), which aims to map the source and target domains to the same distribution, is one of the most popular approaches and has been extensively applied in Wi-Fi sensing \cite{MMD1,MMD2,MMD3,MMD4,MMD5,SN-MMD,airfi}. However, as shown in Fig. \ref{fig:performance}, the global alignment of traditional DAL methods can lead to incorrect alignments within categories, which can hinder the effectiveness of the classification boundary trained in the source domain when applied to the target domain.

Additionally, several cross-domain Wi-Fi sensing methods focus on few-shot learning (feature learning) \cite{wang2025survey}. These methods typically follow a paradigm where a feature extractor is first trained, and then similarities of the extracted features from the source and target domains are compared for classification. Most of these methods concentrate on efficiently designing and improving the feature extractor and similarity calculation components.
For instance, Ding et al. \cite{ding2021wi} propose using a CNN trained in the source domain as a feature extractor, employing cosine similarity in the feature embedding for similarity calculations. They also implement a low-pass Butterworth filter for denoising. Building on this, AutoFi \cite{AutoFi} is introduced (without the low-pass denoising), proposing a pre-training method to enhance the feature extractor's capacity within the neural network.
In Yang et al. \cite{yang2019learning}, a different similarity calculation method is employed, specifically a Siamese network \cite{siamese}. They utilize a CNN combined with Bi-LSTM as the feature extractor. Expanding upon this, CrossFi \cite{CSi-Net} uses a ResNet-based Siamese architecture, introducing a template generation method to enhance feature representation and an attention-based similarity calculation method to better capture relationships between samples. In CrossFi, the application of the Siamese network is extended to a zero-shot scenario.
However, a widely recognized issue in few-shot learning methods is low stability: the model's performance on the testing set can be unstable during training (as referenced in \cite{CSi-Net} and Section \ref{Comparison with Traditional DA Methods}), with accuracy varying sharply between successive epochs. This instability suggests that simply stopping the model training at a fixed epoch may result in poor performance. If we wish to implement an early stopping method, we require labeled validation data from the target domain. However, the only labeled data available from the target domain is the support set, which is often a subset of the training set and cannot be used as a validation set.

\color{black}

\subsection{Our Contributions}
To address the limitations of traditional DAL methods in cross-domain Wi-Fi sensing, we propose a novel Few-Shot Learning (FSL) framework, called K-Nearest Neighbors Maximum Mean Discrepancy (KNN-MMD). Our approach introduces a local distribution alignment strategy that aligns the source and target domains within each category, rather than globally aligning their overall data distributions. This refinement enhances the effectiveness of traditional DAL techniques by addressing inter-category misalignments. Additionally, our framework ensures that the support set is excluded from the network’s training process, enabling the implementation of an early stopping strategy based on support set performance. This ensures stable and consistent target domain performance while avoiding the challenges faced by traditional DAL methods in determining an optimal stopping point. The key contributions of this work are summarized as follows:




\noindent \textbf{1. A Few-Shot Learning Framework with Local Alignment:}
We present KNN-MMD, a novel FSL framework tailored for cross-domain Wi-Fi sensing. Unlike traditional DAL methods that rely on global alignment, our framework leverages a category-specific alignment strategy to improve accuracy and robustness. Specifically, we use K-Nearest Neighbors (KNN) to assign pseudo-labels to the target domain and construct a high-confidence ``help set." This help set, in combination with the source domain training data, is then used to perform local alignment during the training of the classification network.

\noindent \textbf{2. An Early Stopping Strategy Using the Support Set:}
Our framework introduces a practical early stopping mechanism by excluding the support set from the training process while using it as a validation set for monitoring accuracy and loss. This approach ensures the training process concludes at an optimal point, maintaining strong performance in the target domain. Traditional DAL methods often lack such a mechanism, leading to unstable or suboptimal performance.

\noindent \textbf{3. Comprehensive Experimental Validation:}
We extensively evaluate our method across multiple Wi-Fi sensing tasks, including gesture recognition, person identification, fall detection, and action recognition. Experimental results demonstrate that our framework outperforms existing models in both accuracy and stability, showcasing its effectiveness in real-world cross-domain Wi-Fi sensing scenarios.

The structure of this paper is as follows. In Section \ref{Preliminary}, we first introduce the Wi-Fi sensing system and analyze the cross-domain Wi-Fi sensing tasks. In Section \ref{Domain Adaptation Methods}, we review traditional DA methods across different types and illustrates their limitations. In Section \ref{Methodology}, we revisit the cross-domain task and propose our local alignment method in detail. In Section \ref{Experiment}, we use a series of experiments to demonstrate the efficiency and stability of our method. Finally, in Section \ref{Conclusion}, we conclude the paper and provide some points for future research.

\section{System Model and Problem Statement} \label{Preliminary}
\subsection{Wi-Fi Sensing Principle}

In Wi-Fi sensing, CSI is one of the most commonly used features. The channel between the transmitter and receiver can be influenced when encountering dynamic or static objects. The {\color{black} receiver (RX)} will then receive the multipath superimposed signals from refraction, reflection, and scattering. By analyzing the received CSI, we can obtain information about the environment and behavior. The channel model can be mathematically represented as:
\begin{equation}
Y=HX+N,
\end{equation}
where $Y$ and $X$ are the matrices of the received and transmitted signals, respectively, $N$ is the noise, and $H$ is the estimate of the Channel Frequency Response (CFR) of the wireless channel, which contains information about the amplitudes, phases, and delays of the multipath components. It can also be represented as:
\begin{equation}
H=||H||e^{j \angle H}
\end{equation}
where $||H||$ and $\angle H$ denote the amplitude and phase of the CSI measurement, respectively.

For most Wi-Fi sensing tasks, when an object moves, the CSI changes according to patterns specific to the movement. By analyzing these changes in CSI values, the object motions can be detected. Compared to other Wi-Fi signals, such as the Received Signal Strength Indicator (RSSI), CSI captures fine-grained features, thus providing more precise information for Wi-Fi sensing.

\subsection{Domain Shift Issue in Wi-Fi Sensing}


\begin{figure}[htbp]
\centering 
\includegraphics[width=0.4\textwidth]{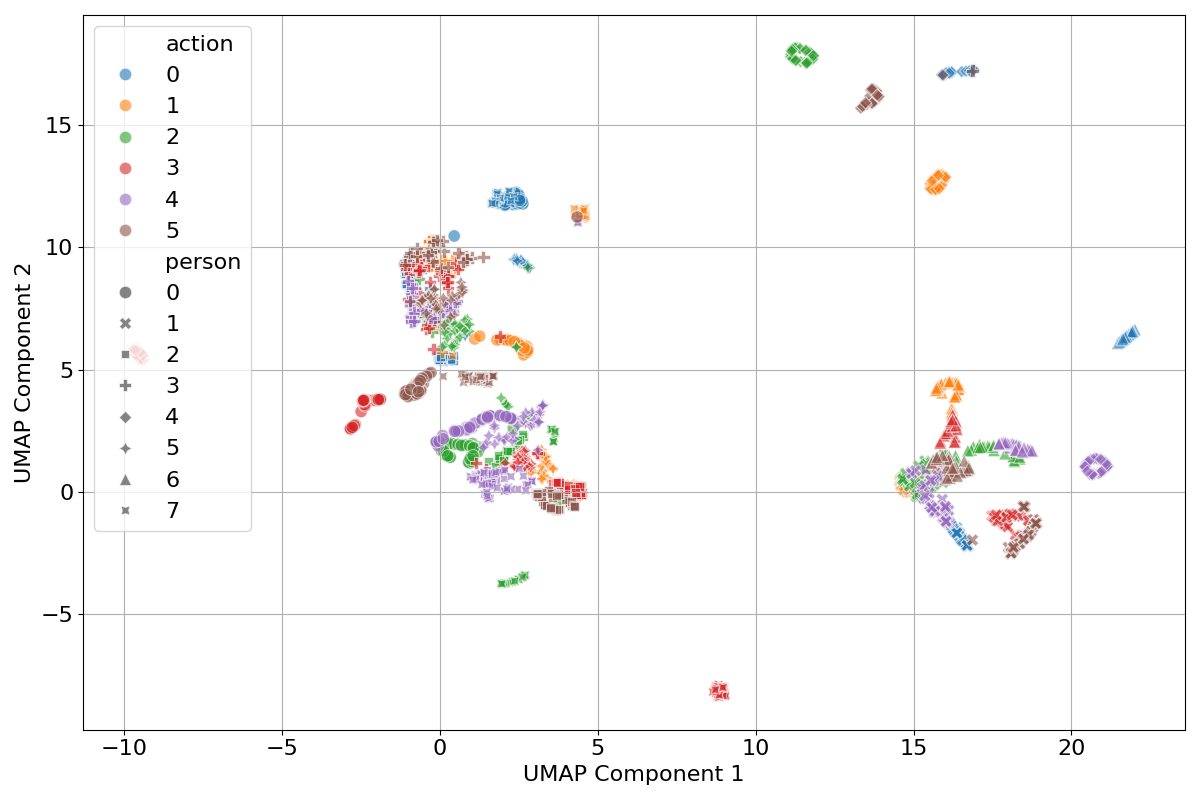}
\caption{\textcolor{black}{UMAP Dimensionality Reduction of the WiGesture Dataset \cite{CSI-BERT}: In the figure, different shapes represent different person IDs, while different colors represent different action IDs. The x and y axes correspond to the two main UMAP components.}}
\label{fig:Wigesture Analysis}
\end{figure}


To facilitate understanding, we first define the symbols and the cross-domain scenarios. In the following, the variables $x$ and $y$ refer to the input data and output label, respectively. $\theta$ represents the network parameters. $P_s$ and $P_t$ represent the classification probabilities of each category in the source domain and target domain, respectively. Furthermore, the focus of this paper is primarily on the few-shot cross-domain task. Most available data is in the source domain {\color{black} except} $n$ labeled samples for each category in the target domain. This scenario is often referred to as the n-shot problem. In addition, we also have some unlabeled data from the target domain. Our objective is to accurately classify these unlabeled data. For the sake of simplicity, we will refer to the labeled data from the source domain as the training set, and the labeled data from the target domain as the support set.

\begin{figure}[htbp]
\centering 
\subfloat[People ID 0]{\includegraphics[width=0.2\textwidth]{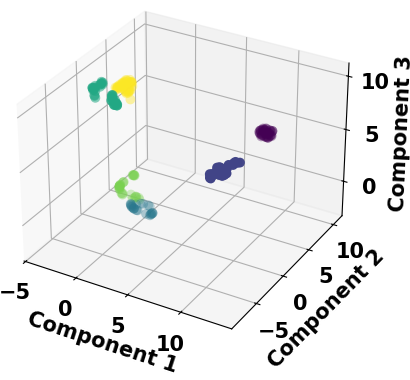}\label{source}} 
\subfloat[People ID 0-7]{\includegraphics[width=0.2\textwidth]{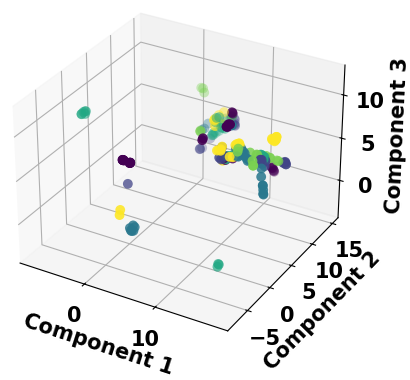}\label{target}} \\
\subfloat[ResNet Embedding Result]{\includegraphics[width=0.4\textwidth]{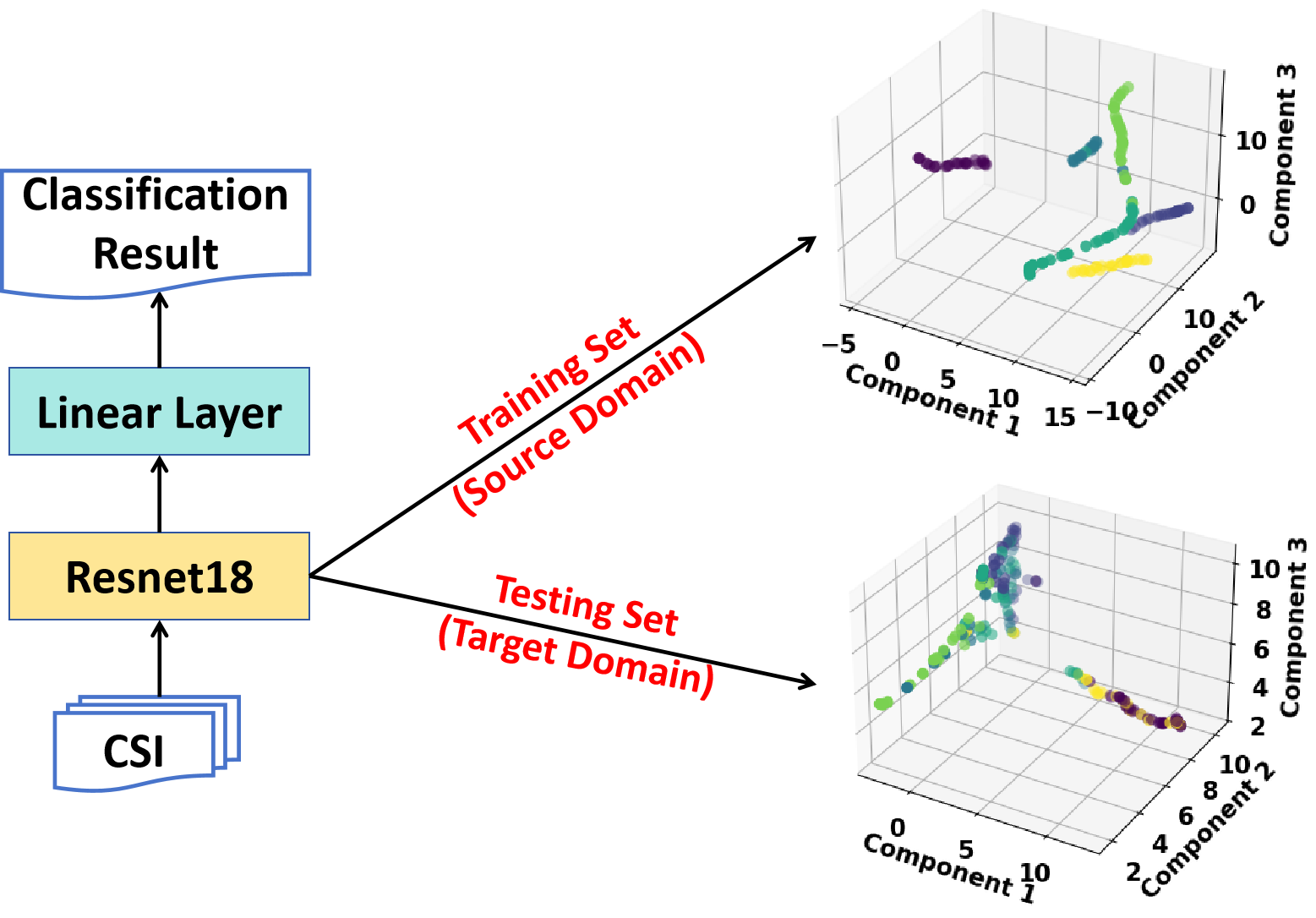}\label{ResNet}}
\caption{\color{black}{Data Dimension Reduction Results of the WiGesture Dataset \cite{CSI-BERT}: Different colors represent different action categories. The data dimensions have been reduced to three for visualization, corresponding to the three axes in the figures. They remain consistent across subsequent figures.}}
\label{fig:domain shift}
\end{figure}

In cross-domain Wi-Fi sensing, the most significant challenge is the substantial variation in data distribution across different domains, which causes models trained in the source domain to often fail in the target domain. For better illustration, we use WiGesture dataset \cite{CSI-BERT}, which contains the CSI collected from 8 individuals performing 6 different gesture actions, to illustrate the domain shift problem in detail. For the gesture recognition or action classification task, we take different people as different domains. And for the people identification task, we take different actions as different domains. 
\color{black}
Fig. \ref{fig:Wigesture Analysis} depicts the Uniform Manifold Approximation and Projection (UMAP) based data reduction \cite{UMAP} (this method will be introduced in Section \ref{Preliminary Classification}). It is evident that even for the same action, individuals exhibit distinct data distributions; similarly, for the same individual, different actions show varied data distributions. These significant domain gaps can pose challenges for classification models. 
To illustrate this more clearly, Figs. \ref{source} and \ref{target} display the UMAP results for single and multiple individuals. It is apparent that within a single individual, different actions exhibit significant dissimilarities. However, when considering different individuals, the boundaries separating different actions become less distinct. 
\color{black}
These findings imply that it is challenging to generalize the learned features from one set of individuals to others. In Fig. \ref{ResNet}, a ResNet18 \cite{ResNet} is trained by setting the  people with IDs 1-7 as the source domain and people with ID 0 as the target domain. It can be observed that the trained ResNet18 can distinguish different actions in the source domain but fail to distinguish different actions in the target domain. This is mainly because even for the same action, different people have distinct behavioral patterns, which can lead to significant gaps in the CSI patterns between individuals. As a result, to make the model perform well in the target domain, it is necessary to address the domain shift (cross-domain) problem.

\section{Preliminary} \label{Domain Adaptation Methods}

\begin{table*}
\caption{Comparison of Different DA Methods}
\centering
\begin{adjustbox}{width=1.00\textwidth}
\begin{tabular}{|c||c|c|c||c|}
\hline
 & \textbf{Metric-based Method} & \textbf{Learning-based Method} & \textbf{Domain Alignment Method} & \textbf{Ours}\\
\hline
\textbf{Representative Methods} & KNN\cite{KNN}, K-means\cite{kmeans} & Siamese\cite{siamese}, Triplet Network\cite{triplet}  & MMD\cite{MMD}, GFK\cite{GFK,zhang2018grassmannian} & KNN-MMD\\
\hline
\textbf{Sensitivity to Quality of Support Set} & High & Moderate & \textbf{None} & Low \\
\hline
\textbf{Stability} & Low & Low & Low & \textbf{High} \\
\hline
\textbf{Assumption $P_t(y|x) = P_s(y|x)$} & \textbf{No} &  Some methods require it\cite{DANN} & Yes & \textbf{No} \\
\hline
\end{tabular}
\end{adjustbox}
\label{DA Method}
\end{table*}


As the most prevalent method in cross-domain tasks, DA encompasses various classification approaches. We primarily follow \cite{zhu2024cross, he2024domain}, categorizing DA into three types: metric-based methods, learning-based methods, and DAL methods. In practice, a method may belong to multiple categories, however, this paper focuses on their primary classifications: metric-based methods refer mainly to traditional machine learning approaches that do not involve training, learning-based methods pertain to those associated with novel network architectures, and DAL methods primarily address the alignment of feature spaces between source and target domains.

The core idea of metric-based methods is to learn a good sample representation and metric. However, most metric-based methods only depend on the support set without using the training set in source domain sufficiently. As a result, these methods are heavily influenced by the quality of the support set, which can lead to unstable performance and the waste of the training set.

With the development of deep learning, there have been many learning-based methods for cross-domain tasks in recent years, including transfer learning \cite{transfer}, meta learning \cite{meta}, FSL \cite{few_shot}. However, an important practical problem of these methods is that we do not know when to stop the model training. Since there is often a significant gap between the source and target domains, we cannot reliably infer the model's performance on the target domain based on its performance on the source domain. As a result, the typical early stop strategies cannot be easily applied in such scenarios.


\color{black}
DAL methods aim to map the source domain and target domain to a common feature distribution, allowing the data in both domains to share the same distribution in the new space. However, most DAL methods, such as MMD \cite{MMD}, Kernel Mean Matching (KMM) \cite{KMM}, and Subspace Geodesic Flow (SGF) \cite{SGF}, rely on the assumption that the conditional probabilities in the source and target domains are the same, i.e., $P_t(y|x) = P_s(y|x) = P(y|x)$, where $P_s$ and $P_t$ represent the probability distributions in the source and target domains, respectively, and $x$ and $y$ represent the data sample and ground truth category label.
When training the network in the source domain, the network parameters $\theta$ need to be learned to achieve $P(y|x;\theta) \approx P(y|x)$ under the source domain $P_s(x)$. However, since $P_t(x) \neq P_s(x)$, the $P(y|x;\theta)$ learned from the source domain cannot generalize to the target domain $P_t(x)$. Consequently, these traditional methods can be viewed as attempting to find a latent space where the input data from the source and target domains are mapped to the same distribution, i.e., $P_t(\text{g}(x)) \approx P_s(\text{g}(x))$, where $\text{g}(\cdot)$ is a mapping function. This allows the model to learn the parameters $\theta$ more effectively and accurately in the target domain.
However, in many practical scenarios, the assumption $P_t(y|x) = P_s(y|x)$ is not satisfied. This has been demonstrated in our experiments (Section \ref{Experiment Result}) as well as in our previous research \cite{CSi-Net}, where traditional DAL methods have even failed in cross-domain Wi-Fi sensing tasks. A more detailed derivation of the formulas can be found in Section \ref{Local Distribution Alignment}.
\color{black}

In conclusion, the quality of the support set makes the performance of metric-based methods unstable. For learning-based methods, they also use source domain data to train, and the influence of the support set is not as significant as for metric-based methods. However, it's hard to identify when to stop the training process, and the accuracy in the test set always has significant fluctuations during training (shown in Section \ref{Experiment Result}). Finally, most DAL methods do not require a support set for training, and they are not affected by the quality of the support set. But they mainly rely on the assumption of $P_t(y|x) = P_s(y|x)$, which is not always established in practice. Thus, the model would easily fail.

In addition, some other researches have also identified the limitations of traditional DAL methods. Tian et al. \cite{KNN-MMD} utilize KNN \cite{KNN} to identify the nearest samples in the entire data domain to each sample in the source domain, expanding the distance between neighboring samples from different categories in the source domain, and shrinking the distance between neighboring samples from the target domain under the assumption that such sample pairs share the same category. However, this approach may also lead to potential failures, as shown in Fig. \ref{fig:performance}. Though the source domain samples have a relatively small distance to their neighboring samples in the target domain and a large distance to other category samples in the source domain, it cannot guarantee accurate classification in the target domain. Li et al.\cite{EEG} first generate pseudo-labels for samples in the target domain using a classifier trained in the source domain and then attempt to expand the distance between samples with different categories and shrink the distance between samples with the same category. However, due to the inherent gap between the source and target domains, the pseudo-labels generated in the first step may be inaccurate, which can adversely impact the subsequent steps.

\begin{figure}[htbp]
\centering 
\subfloat[Confusion Matrix]{\includegraphics[width=0.4\textwidth]{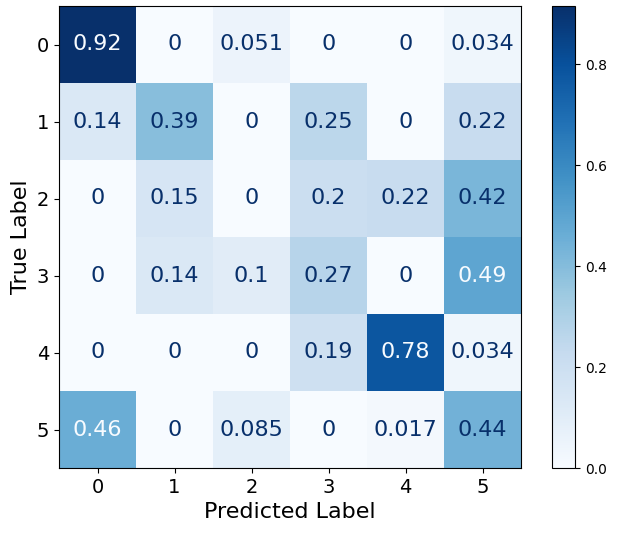}\label{fig:confusion}} \\
\subfloat[Accuracy under Top $p\%$ Confidence Level]{\includegraphics[width=0.4\textwidth]{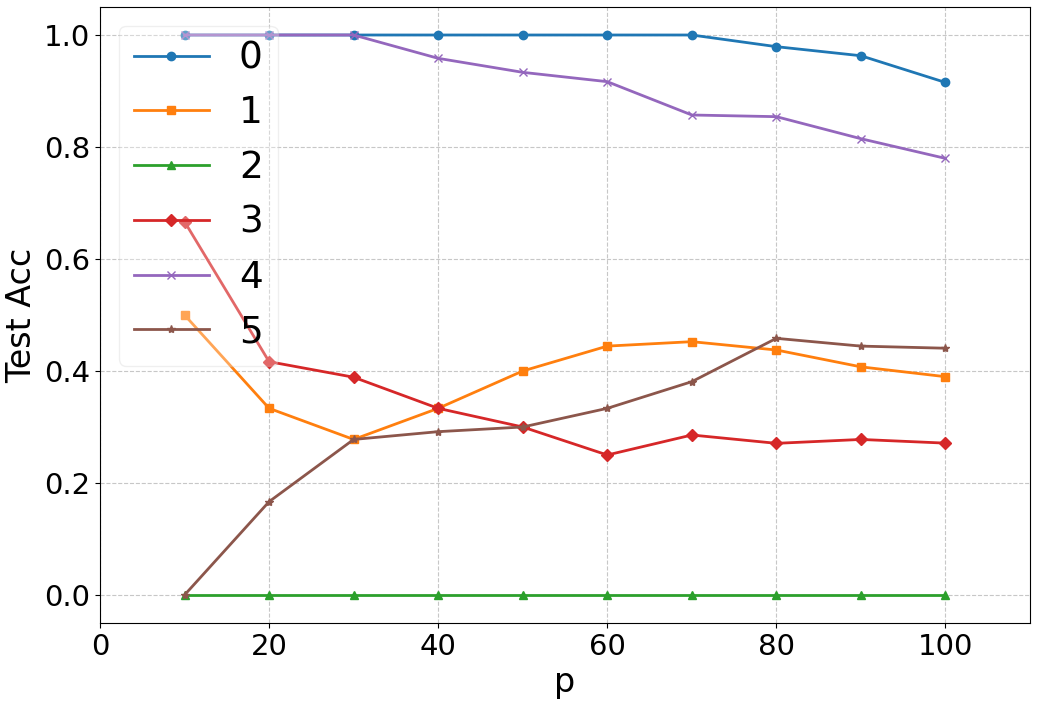}\label{fig:p_vanilla}} \\
\caption{\color{black}{Model Performance of Vanilla ResNet18 on WiGesture Dataset}}
\label{fig:Vanilla}
\end{figure}

\color{black}
\begin{remark}
Although some previous work has highlighted the disadvantages of conventional global alignment and proposed similar pseudo-label-based DAL methods, we want to emphasize the differences between our work and theirs. First, we argue that some prior DAL methods still struggle in challenging scenarios. For instance, the method proposed by \cite{KNN-MMD} cannot effectively address the scenario illustrated in Fig. \ref{fig:performance}, as mentioned in the previous section. Our proposed method, which directly aligns the source and target domains within each category, represents a simple yet effective approach to achieve local alignment. The success of our method primarily relies on the quality of pseudo labels for the target domain. An inevitable issue in pseudo-label-based methods is the occurrence of incorrect classifications and noise in the data. Most previous work \cite{EEG, ge2023unsupervised, zhu2020deep} first trains a network in the source domain and then uses it to classify the target domain, generating pseudo labels. Typically, these methods retain only the top classification samples from the target domain for subsequent DAL training, aiming to minimize noise and incorrect primary classifications as much as possible. While these methods have shown considerable success in fields like CV and electrocardiogram (ECG) analysis, the Wi-Fi CSI serves as a different modality, often exhibiting a completely different data distribution between source and target domains. This discrepancy can render the network trained on the source domain ineffective in the target domain. For example, using the WiGesture dataset, we train a ResNet18 in the source domain and test its performance in the target domain. The confusion matrix is shown in Fig. \ref{fig:confusion}. For category 2-4, most classification results are incorrect. Notably, in category 2, there are no correct generated labels, which hinders subsequent operations and leads to significantly low model performance. In Fig. \ref{fig:p_vanilla}, we also display the accuracy across different categories under the top \(p\%\) classification confidence level. For classes other than 0 and 4, regardless of how \(p\%\) is selected, the accuracy remains low, resulting in poor quality of pseudo labels. In contrast, our method utilizes model-based KNN for pseudo label generation, which is less affected by cross-domain issues compared to neural networks. Additionally, we propose a confidence-level-based approach to mitigate the influence of noise and incorrect classifications, which has been demonstrated to be effective through sensitivity analysis. Furthermore, our method supports an early stopping mechanism, which is not feasible for most DAL methods. This feature allows us to efficiently determine when to stop the model at the optimal point, significantly enhancing model stability.
\end{remark}
\color{black}

\section{Methodology} \label{Methodology}
\begin{figure*}
\centering 
\includegraphics[width=0.8\textwidth]{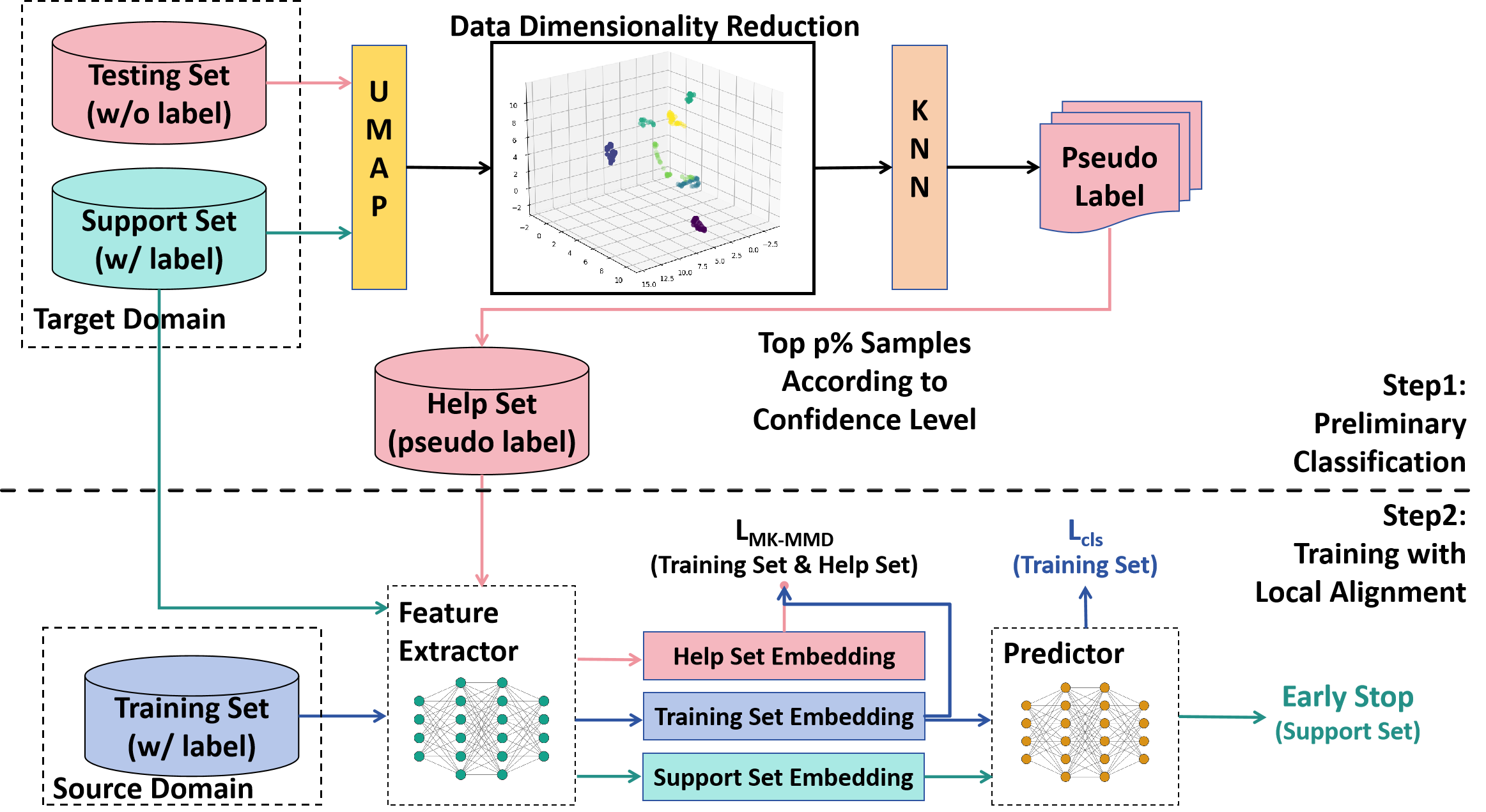}
\caption{Workflow: Our method primarily consists of two steps. In Step 1, we use KNN to preliminarily classify the testing set and select the top $p\%$ of samples with the highest classification confidence levels to construct a help set. In Step 2, we train a network for the final classification, using local alignment based on the training set and the help set. Simultaneously, we employ an early stop strategy according to the classification accuracy in the support set.}
\label{fig:workflow}
\end{figure*}

\begin{figure}[htbp]
\centering 
\includegraphics[width=0.4\textwidth]{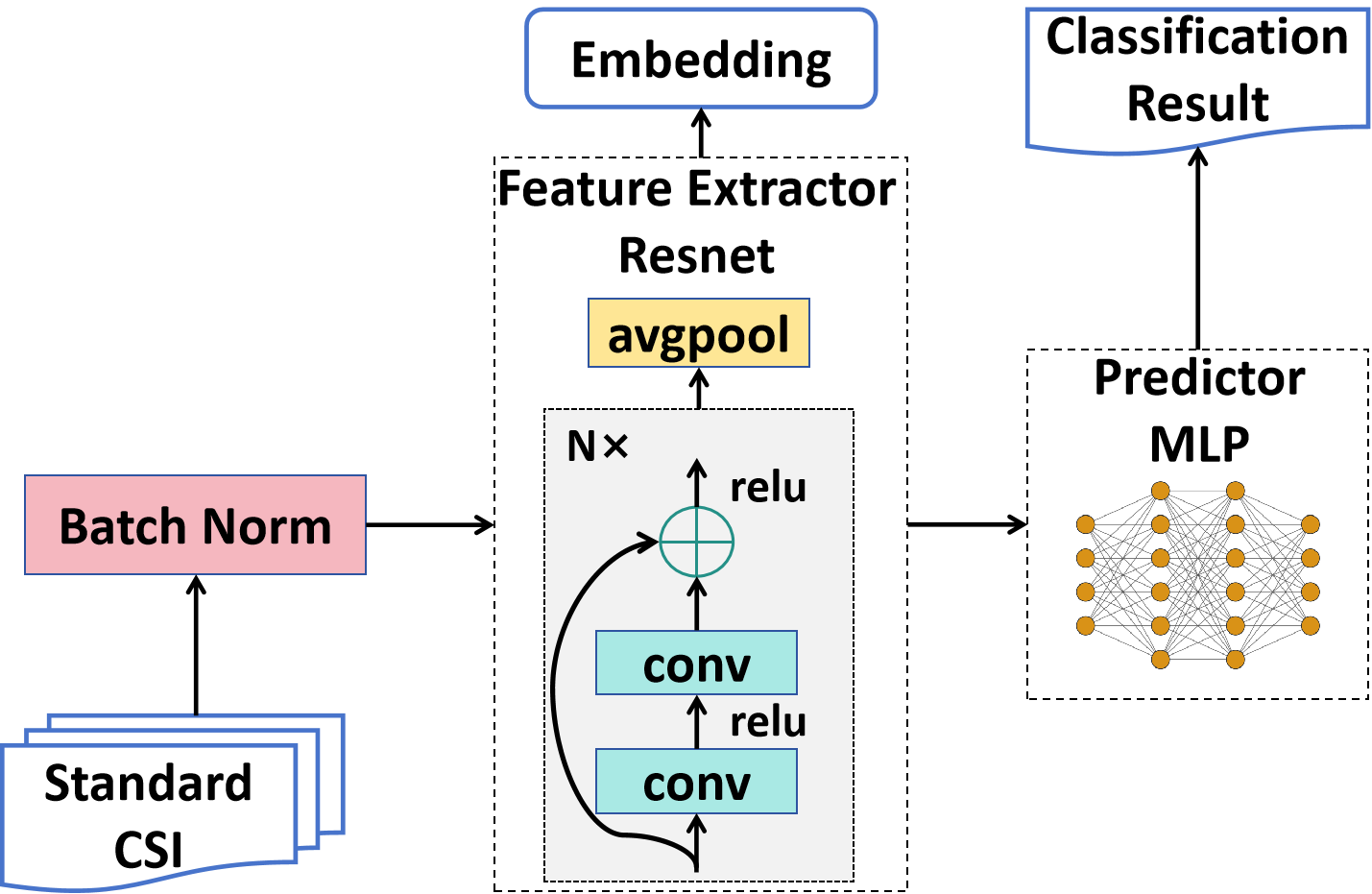}
\caption{Model Structure: We use the ResNet18 \cite{ResNet} as the feature extractor and a three-layer MLP as the predictor.}
\label{fig:network}
\end{figure}

\subsection{Overview}


As discussed in previous section, traditional DAL methods cannot guarantee model performance. To solve this problem, we propose a local distribution alignment approach that aligns the distribution between the source domain and the target domain within each category. (The detailed mathematical analysis will be discussed in Section \ref{Local Distribution Alignment}.) However, the category labels in the testing set cannot be accessed during training, and it is also challenging to estimate the target domain distribution solely based on the limited labeled data in the support set. As a result, we propose a preliminary strategy to first obtain some labels in the testing set to estimate the distribution.

As shown in Fig. \ref{fig:workflow}, our method mainly has two steps: (1) Preliminary Classification and (2) Training Network with Local Alignment. In Step 1, we use KNN to preliminarily classify the testing set according to the support set. We then choose the samples with the highest classification confidence levels to construct a help set for Step 2. The help set will be used for the local alignment with the training set. In Step 2, we train a classification network with the Multiple Kernel Maximum Mean Discrepancy (MK-MMD) loss function to align the source domain and the target domain within each category. Additionally, we use an early stop strategy according to the classification accuracy in the support set, as it does not participate in the training and has the same distribution as the testing set. By this way, our training process can stop at a point where the testing set has a relatively high classification accuracy.

Before introducing our method in detail, let us describe the data format in our study. Each sample consists of two dimensions representing the data length $l$ and subcarrier $s$, respectively. As we choose ResNet as the feature extractor, it requires an extra channel. Consequently, each sample has three dimensions, denoted as $x \in \mathbb{R}^{1 \times l \times s}$. The training set, support set, help set, and testing set are represented as $X^{train}$, $X^{support}$, $X^{help}$, and $X^{test}$, respectively, where the help set is not provided but generated by us.

\subsection{Local Distribution Alignment} \label{Local Distribution Alignment}

\color{black}
Let's reconsider the cross-domain classification task. During training, we have a labeled dataset from the source domain, denoted as $D_s = (X_s, Y_s)$, and an unlabeled dataset from the target domain, represented as $D_t = X_t$. Here, $X_s$ and $X_t$ are the sample data from the source and target domains, respectively, while $Y_s$ and $Y_t$ denote their category labels. When we train a network using $D_s$, we obtain the network parameters $\theta$ through Maximum Likelihood Estimation (MLE):
\begin{equation}
\theta = \arg \max_\theta P(Y_s | \text{f}(X_s; \theta)) \ ,
\end{equation}
where $\text{f}(\cdot; \theta)$ represents the neural network. However, due to the significant domain gap between the source and target domains, the network cannot guarantee the probability of samples in the target domain, i.e., $P(Y_t | \text{f}(X_t, \theta))$. In conventional DAL methods, a mapping function $\text{g}(\cdot)$ is introduced to align the data from the source and target domains to a similar distribution, such that $P(\text{g}(X_s)) \approx P(\text{g}(X_t))$, which we refer to as global alignment.

However, we note that this approach does not ensure model performance in the target domain. Consequently, the network learns the parameters $\theta$ as follows:
\begin{equation}
\theta = \arg \max_\theta P(Y_s | \text{f}(\text{g}(X_s); \theta)) \ .
\end{equation}
This can be interpreted as the network fitting the label distribution in the source domain:
\begin{equation}
P(y | \text{f}(\text{g}(x); \theta)) \approx P_s(y | \text{g}(x)) \ ,
\end{equation}
where $x$ and $y$ are arbitrary data samples and their corresponding ground truth labels, respectively. The distributions $P_s$ and $P_t$ represent the probabilities in the source and target domains, respectively. For the network to function effectively in the target domain, we require that $P_s(y | \text{g}(x)) \approx P_t(y | \text{g}(x))$. According to Bayes' rule, we have:
\begin{equation}
\begin{aligned}
P_s(y | \text{g}(x)) &= P_t(y | \text{g}(x)) \frac{P_s(\text{g}(x) | y) P_s(y)}{P_s(\text{g}(x))} \frac{P_t(\text{g}(x))}{P_t(\text{g}(x) | y) P_t(y)} \\
&= P_t(y | \text{g}(x)) \frac{P_t(\text{g}(x))}{P_s(\text{g}(x))} \frac{P_s(\text{g}(x) | y)}{P_t(\text{g}(x) | y)} \frac{P_s(y)}{P_t(y)} \\
&= P_t(y | \text{g}(x)) \frac{\sum_{y' \in \tilde{Y}} P_t(y') P_t(\text{g}(x) | y')}{\sum_{y' \in \tilde{Y}} P_s(y') P_s(\text{g}(x) | y')} \\
& \quad \cdot \frac{P_s(\text{g}(x) | y)}{P_t(\text{g}(x) | y)} \frac{P_s(y)}{P_t(y)} \ ,
\label{MMD problem}
\end{aligned}
\end{equation}
where $\tilde{Y}$ denotes the set of all categories. In global alignment, we only have $P_s(\text{g}(x)) \approx P_t(\text{g}(x))$. However, this alone is insufficient to ensure $P_s(y | \text{g}(x)) \approx P_t(y | \text{g}(x))$. In most cases, we find that $P_s(y) \approx P_t(y)$, indicating that the label distributions are similar between the source and target domains. If we also have $P_t(\text{g}(x) | y) \approx P_s(\text{g}(x) | y)$, then it follows that $P_s(y | \text{g}(x)) \approx P_t(y | \text{g}(x))$. We refer to this sufficient condition as local alignment.

\color{black}

To realize distribution alignment, in this paper, we utilize the MK-MMD method, which is based on MMD. MMD measures the distance between two distributions and aims to map the source and target domains to a middle space with similar distributions.

If two distributions have exactly the same $k$-th central factorial moment for any $k$, then they are regarded following the same distribution. Since the $k$-th central factorial moment can be represented by the $1$-st central factorial moment, i.e. the mean value, in a feature space, we can search for a mapping feature that maximizes the difference in means between the source and target domains. By minimizing the maximum distance, we can minimize the distribution difference between the two domains. Accordingly, MMD is defined as:
\begin{equation}
\begin{aligned}
\text{MMD}[F,p,q]:=\text{sup}_{f \in F}|\text{E}_p[f(x)]-\text{E}_q[f(x)]| \ , 
\label{MMD}
\end{aligned}
\end{equation}
where $F$ is the set of mapping functions in the Reproducing Kernel Hilbert Space (RKHS), $p$ and $q$ represent two distributions, and $\text{E}$ denotes the expectation value. However, directly computing the mapping function can be challenging in practice. Therefore, kernel functions are often used to compute MMD as:
\begin{equation}
\begin{aligned}
& \text{MMD}[F,p,q] =  \text{sup}_{f \in F} [|\frac{1}{n}\sum_{i=1}^nf(x_i^{(p)})-\frac{1}{m}\sum_{i=1}^mf(x_i^{(q)})|^2]^{\frac{1}{2}} \\
&  =  \text{sup}_{f \in F} [\frac{1}{n^2}\sum_{i=1}^n\sum_{j=1}^nf(x_i^{(p)})f(x_j^{(p)}) \\
& -  \frac{2}{nm}\sum_{i=1}^n\sum_{j=1}^mf(x_i^{(p)})f(x_j^{(q)})  +  \frac{1}{m^2}\sum_{i=1}^m\sum_{j=1}^mf(x_i^{(q)})f(x_j^{(q)})]^{\frac{1}{2}} \\
&  =  \text{sup}_{k} [\frac{1}{n^2}\sum_{i=1}^n\sum_{j=1}^nk(x_i^{(p)},x_j^{(p)}) \\
& - \frac{2}{nm}\sum_{i=1}^n\sum_{j=1}^mk(x_i^{(p)},x_j^{(q)}) + \frac{1}{m^2}\sum_{i=1}^m\sum_{j=1}^mk(x_i^{(q)},x_j^{(q)})]^{\frac{1}{2}} \ ,
\label{MMD kernal}
\end{aligned}
\end{equation}
where $x_i^{(p)}$ represents the $i$-th sample from distribution $p$, $n$ and $m$ represent the sample sizes of distributions $p$ and $q$ respectively, and $k$ denotes the kernel function of the mapping function $f$.

However, it is often difficult to find a kernel function that can realize the upper bound in practice. Therefore, Gretton et al. \cite{MK-MMD} proposed MK-MMD, which uses multiple kernel functions to approximate the upper bound, as:
\begin{equation}
\begin{aligned}
& \text{MK-MMD}^2[K,p,q] = \sum_{h=1}^H \beta_h[\frac{1}{n^2}\sum_{i=1}^n\sum_{j=1}^nK_h(x_i^{(p)},x_j^{(p)}) \\
& - \frac{2}{nm}\sum_{i=1}^n\sum_{j=1}^mK_h(x_i^{(p)},x_j^{(q)}) + \frac{1}{m^2}\sum_{i=1}^m\sum_{j=1}^mK_h(x_i^{(q)},x_j^{(q)})] \ , 
\label{MK-MMD}
\end{aligned}
\end{equation}
where $K$ is the set of kernel functions provided by the user, $\beta$ is a set of weights that satisfy $\sum_{h=1}^H \beta_h=1$, and $H$ is the kernel amount. During training, the MK-MMD loss function can be used in the embedding result of the source and target domains to encourage the feature extractor to map the source and target domain data to the same distribution in the intermediate feature space.

\subsection{Model Structure}

\color{black}

Our model architecture is illustrated in Fig. \ref{fig:network}, which primarily consists of three components: a batch normalization layer, a feature extractor, and a final classifier. The model takes the standard CSI sequence as input, where the signal is first standardized over the time dimension, a method shown to be effective for CSI sensing model training \cite{CSI-BERT}. In the network, the batch normalization layer helps reduce the domain gap between different samples, facilitating the training process. We then select the pre-trained ResNet18 \cite{ResNet} as the feature extractor, as the Wi-Fi signal spectrum exhibits high similarity to images. Convolutional networks, which effectively extract local features, have demonstrated excellent performance in CSI classification \cite{CSi-Net, fall, cai2023falldewideo}. Furthermore, inspired by transfer learning, the pre-trained parameters from image data provide a strong starting point for Wi-Fi sensing \cite{zhu2024srcsense}. Finally, we employ a multilayer perceptron (MLP) to process the embeddings for classification. The detailed description is as follows:

\color{black}

After batch normalization, the processed data is fed into ResNet to obtain the embedding, which serves as the feature space mapping mentioned in Section \ref{Local Distribution Alignment}. Finally, the MLP acts as a classifier, taking the embedding as input and producing the corresponding label as output. \textcolor{black}{The network can be expressed as:}
\begin{equation}
\begin{aligned}
E&=\text{ResNet}(\text{BatchNorm}(X)) \ , \\
\hat{Y}&=\text{MLP}(E) \ ,
\label{eq:network}
\end{aligned}
\end{equation}
where $X$ is the input, $E$ is the embedding result, and $\hat{Y}$ is the output.

\subsection{Training Process} \label{Workflow}
The training process consists of two main phases: preliminary classification and network training with local alignment. In order to align the source and target domains locally within each category, we need to know the labels of the target domain samples. However, since we have only a very limited labeled support set in the target domain, it is challenging to use these samples for alignment, as this typically requires a larger number of samples to capture statistical features effectively. To address this issue, we propose using KNN for preliminary classification of the target domain. Although KNN may not be very stable and often produces numerous incorrect classifications, we find that classification accuracy is closely related to the confidence level. By leveraging this insight, we can construct a helper set consisting of samples with high confidence levels, the majority of which are correctly classified in practice. During the network training phase, we use this helper set to achieve local alignment with the source domain training set. Finally, the trained network can be employed for final classification, demonstrating excellent performance on samples with low classification confidence from KNN. Moreover, since the support set does not participate in network training, we can utilize it as a validation set to determine when to stop model training. In the following, we provide a detailed description of the two phases:

\subsubsection{Preliminary Classification} \label{Preliminary Classification}
Before applying KNN for preliminary classification, we need to flatten each sample to one-dimensional data, denoted as $x' \in R^{ls}$. Subsequently, we construct the target domain data $X'^{target}=[X'^{support},X'^{test}]$. However, the data dimension $ls$ is often too high, exceeding a thousand, which may lead to a high computational burden for KNN. Therefore, prior to using KNN, we aim to reduce the data dimension in $X'^{target}$ via UMAP \cite{UMAP}.


UMAP primarily involves four steps: constructing the nearest neighbor graph, approximating a fuzzy-simplicial set, optimizing the low-dimensional embedding using stochastic gradient descent, and finalizing the embedding after convergence. Since UMAP is not the primary focus of our research, we directly utilize the function implemented by the UMAP package in Python. For more details, interested readers can refer to the UMAP paper \cite{UMAP}.

After performing dimension reduction, we utilize the KNN algorithm to classify the testing set based on the support set. For more detailed information on the KNN algorithm, please refer to the paper \cite{KNN}. Briefly, for each sample in the testing set, we identify the nearest $k$ samples from the support set using {\color{black}Euclidean distance} as the distance measure. Subsequently, we determine the most frequent label among the identified neighbors as the classification result. In the case of a tie, we select the label of the nearest neighbor as the classification result.

\color{black}
Next, we select the samples from the testing set that have the top $p\%$ confidence level of classification to construct the help set. In our approach, confidence is measured by calculating the distance between each sample in the testing set and its nearest neighbor in the support set with the same label. A smaller distance indicates a higher confidence level. The detailed construction process is outlined in Algorithm \ref{alg:top}.
\color{black}

\begin{algorithm}
\caption{\color{black}{Help Set Construction}}
\label{alg:top}
\textbf{Require}: \\
\hspace*{1em} $[X^{support},Y^{support}]$: Support Set  \\
\hspace*{1em} $[X^{test}]$: Testing Set  \\
\hspace*{1em} $k$: Neighbor Amount in KNN   \\
\hspace*{1em} $p\%$: Use Top-$p\%$ Confidence Testing Samples as Help Set \\
\hspace*{1em} $\epsilon$: Small constant for numerical stability
\begin{algorithmic}[1]
    \State Initialize empty list $\mathit{confidences} \gets \emptyset$
    \For{each sample $x_i \in X^{test}$}
        \State $N \gets \text{KNN}(x_i, X^{support}, k)$ \Comment{Find $k$ nearest neighbors in support set}
        \State $\hat{y}_i \gets \text{majorityVote}(Y^{support}_N)$ \Comment{Predict pseudo-label}
        \State $X^{support}_{\hat{y}_i} \gets \{x_j \in X^{support} \mid Y^{support}_j = \hat{y}_i\}$
        \State $x^* \gets \text{argmin}_{x \in X^{support}_{\hat{y}_i}} \text{distance}(x_i, x)$
        \State $d_i \gets \text{distance}(x_i, x^*)$
        \State $c_i \gets \frac{1}{d_i + \epsilon}$ \Comment{Compute confidence score}
        \State $\mathit{confidences}.\text{append}((x_i, \hat{y}_i, c_i))$ \Comment{Store sample, pseudo-label, and confidence}
    \EndFor
    \State Sort $\mathit{confidences}$ by $c_i$ in descending order
    \State $n \gets \lfloor p\% \times |X^{test}| \rfloor$ \Comment{Number of samples to select}
    \State $\mathit{HelpSet}_X \gets \{x_i \,|\, (x_i, \hat{y}_i, c_i) \in \mathit{confidences}[:n]\}$ \Comment{Selected test samples}
    \State $\mathit{HelpSet}_Y \gets \{\hat{y}_i \,|\, (x_i, \hat{y}_i, c_i) \in \mathit{confidences}[:n]\}$ \Comment{Corresponding pseudo-labels}
    \State \Return $(\mathit{HelpSet}_X, \mathit{HelpSet}_Y)$
\end{algorithmic}
\end{algorithm}

Through experimentation, we have found this method to be effective. For instance, in the four tasks conducted in our experiment, when setting $p\%$ to $50\%$, the classification accuracy of the selected top $p\%$ samples consistently exceeds 90\%, regardless of the values of $k$ (the number of neighbors in KNN) and $n$ (the number of samples per category in the support set for n-shot tasks). Hence, we consider these samples as the help set for the subsequent steps.

\subsubsection{Training Network with Local Alignment}

\textcolor{black}{In Step 2, we train our network using the loss function consisting of three components, presented as:}
\begin{equation}
\begin{aligned}
L = L_{cls}+ \alpha_1 L_{MMD}^{local} + \alpha_2 L_{MMD}^{global} \ ,
\label{loss function}
\end{aligned}
\end{equation}
where $\alpha_1,\alpha_2$ represent the weights. In the loss function, firstly, $L_{cls}$ represents the traditional cross-entropy loss function, which utilizes data solely from the training set. It is used to train the model to learn the basic classification capacity. \textcolor{black}{Secondly, $L_{MMD}^{local}$ represents the local MMD, which calculates the MK-MMD between the help set and the training set within each category:}
\begin{equation}
\begin{aligned}
L_{MMD}^{local} = \frac{1}{M} \sum_{i=1}^M \text{MK-MMD} (K,E^{train},E^{help}|Y=i) \ , 
\label{local mmd}
\end{aligned}
\end{equation}
where {\color{black} $M$ represents the total category amounts}, $E$ is the embedding result of ResNet, $K$ is the given kernel list, and $Y$ is the sample label. It's used to realize our local alignment. \textcolor{black}{Additionally, we add an extra global MMD, which computes the MK-MMD between the entire help set and training set, following the conventional MMD approach, represented as:}
\begin{equation}
\begin{aligned}
L_{MMD}^{global} = \text{MK-MMD} (K,E^{train},E^{help}) \ , 
\label{global mmd}
\end{aligned}
\end{equation}

\color{black}
As mentioned previously, our primary focus lies on the local MMD rather than the global MMD. However, we include the global MMD to address potential challenges arising from an imbalanced distribution of samples among different labels within the help set, as our construction method does not guarantee balanced labels.
Since the MMD is based on the statistical properties of data distribution, a small sample size can introduce random bias. 
By incorporating the global MMD, we can mitigate this issue to some extent. For example, if we ensure $P_s(x|y)=P_t(x|y)$ for all $y$ except $y=i$, and also make $P_t(x)=P_s(x)$, we can achieve $P_s(x|y=i)=P_t(x|y=i)$. 
\color{black}

Nonetheless, it is important to note that this approach has limitations when there is a significant lack of certain categories. Alternatively, we can employ other methods to generate the help set and address category imbalance. For example, within the testing set, we can directly select the top $p\%$ samples within each category or choose a fixed number of samples with the highest confidence level within each category. However, this may impact the accuracy of the help set to some extent.

Furthermore, as previously mentioned, many traditional FSL methods encounter the issue of determining when to stop training. Since their support sets are often involved in the training process, they are unable to obtain a valid set comparable to traditional machine learning methods for early stop. However, in our approach, since the support set is not included in the computation of the loss function, we can implement an early stop strategy based on the accuracy of the support set.

The early stop strategy is outlined in Algorithm \ref{alg:early_stop}. During the training, we keep track of the highest accuracy and the lowest value of the loss function in the support set. Each time the accuracy or the loss value is improved, we save the corresponding model parameters. When the epoch reaches the minimum training epoch $e^{min}$, if the validation accuracy does not improve and the validation loss does not decrease for consecutive $e^{threshold}$ epochs, the model is regarded as converged. Specifically, in our method, when we reach the epoch $e^{min}$, we incorporate two different factors to decrease the historical highest accuracy and increase the historical largest loss. This is because in our experiments, the model sometimes achieves an artificially high accuracy and low loss at the beginning of training, which is not a desirable outcome. \textcolor{black}{The underlying reason for this is the small size of the validation set. Especially in the one-shot scenario, there is only a single sample for each category in the validation set (i.e., the support set), which can lead to relatively high fluctuations in loss and accuracy during training.} Therefore, by utilizing these factors, we aim to increase the likelihood of the neural network preserving the best parameters after reaching $e^{min}$.

\color{black}
\begin{remark}
Here, we want to emphasize the principles and validity of our early stopping approach. During training, we only have labeled training data from the source domain, a small number of labeled support samples from the target domain, and a large amount of unlabeled testing data from the target domain. \textbf{It is important to note that throughout the entire training process, we do not access any ground truth labels for the testing data.} Initially, we obtain pseudo labels for the testing data using KNN by comparing their distances with the labeled support data. We then select a subset of data with the highest confidence levels as the help set, ensuring that the support set is not included in the help set.

Next, we train a neural network using the training data to enhance its classification capacity and utilize both the training data and help data for local alignment. Consequently, during the entire training phase, the network does not access any data from the support set or any ground truth labels from the testing set. Since the support set shares the same distribution as the testing set (both originate from the target domain), we can use the model's performance on the support set as a signal for early stopping. Although the help set is selected using support data, which may introduce some bias, we have found this method to be effective through experimentation. Furthermore, since the ground truth for the testing set remains unavailable throughout the training and validation phases, our method does not risk any potential information leakage.
\end{remark}
\color{black}

\begin{algorithm}
\caption{Training with Adaptive Early Stopping}
\label{alg:early_stop}
\textbf{Require}: \\
\hspace*{1em} $[X^{train},Y^{train}]$: Training data and labels \\
\hspace*{1em} $[X^{support},Y^{support}]$: \textcolor{black}{Support set for validation} \\
\hspace*{1em} $[X^{help},Y^{help}]$: \textcolor{black}{Help set for domain adaptation} \\
\hspace*{1em} $\theta$: Network parameters \\
\hspace*{1em} $K$: Kernel list for MK-MMD \\
\hspace*{1em} $e^{min}$: Minimum training epochs \\
\hspace*{1em} $e^{max}$: Maximum training epochs \\
\hspace*{1em} $e^{threshold}$: Patience threshold for early stopping \\
\hspace*{1em} $\alpha \geq 1$: Loss relaxation factor \\
\hspace*{1em} $\beta \leq 1$: Accuracy relaxation factor 

\begin{algorithmic}[1]
    \State \textbf{Initialize:}
    \State $\text{ResNet}, \text{MLP} \gets$ Initialize network architecture
    \State $loss^{best} \gets \infty$, $acc^{best} \gets 0$,  \Comment{Initialize best validation loss and best validation accuracy}
    \State $e^{acc}, e^{loss} \gets 1$ \Comment{Set epoch counters}
    \State $\theta^{best} \gets \theta$ \Comment{Store best parameters}

    \For{epoch $i = 1$ to $e^{max}$}
        \State \textbf{Forward Pass:}
        \State $E \gets \text{ResNet}(X^{train})$ \Comment{Extract features}
        \State $\hat{Y} \gets \text{MLP}(E)$ \Comment{Compute predictions}
        
        \State \textbf{Loss Computation:}
        \State \textcolor{black}{$l^{train} \gets \mathcal{L}(\hat{Y}, Y^{train};X^{train},X^{support})$} \Comment{Eq. \ref{loss function}}
        \State $\theta \gets \theta - \eta \nabla_\theta l^{train}$ \Comment{Parameter update}
        
        \State \textbf{\textcolor{black}{Validation on Support Set:}}
        \State \textcolor{black}{$l \gets \text{CrossEntropy}(X^{support}, Y^{support})$ }\Comment{Validation loss}
        \State \textcolor{black}{$a \gets \text{Accuracy}(X^{support}, Y^{support})$ }\Comment{Validation accuracy}
        
        \State \textbf{Accuracy Tracking:}
        \If{$a \geq acc^{best}$}
            \State $acc^{best} \gets a$
            \State $e^{acc} \gets 1$
            \State $\theta^{best} \gets \theta$ \Comment{Save improved parameters}
        \Else
            \State $e^{acc} \gets e^{acc} + 1$
        \EndIf
        
        \State \textbf{Loss Tracking:}
        \If{$l \leq loss^{best}$}
            \State $loss^{best} \gets l$
            \State $e^{loss} \gets 1$
            \State $\theta^{best} \gets \theta$ \Comment{Save improved parameters}
        \Else
            \State $e^{loss} \gets e^{loss} + 1$
        \EndIf
        
        \State \textbf{Relaxation at Minimum Epochs:}
        \If{$i == e^{min}$}
            \State $loss^{best} \gets loss^{best} \cdot \alpha$ \Comment{Relax loss threshold}
            \State $acc^{best} \gets acc^{best} \cdot \beta$ \Comment{Relax accuracy threshold}
            \State $e^{acc}, e^{loss} \gets 1$ \Comment{Reset counters}
        \EndIf
        
        \State \textbf{Early Stopping Condition:}
        \If{$i > e^{min}$ \textbf{and} $e^{loss} > e^{threshold}$ \textbf{and} $e^{acc} > e^{threshold}$}
            \State \textbf{break} \Comment{Stop training when both metrics stagnate}
        \EndIf
    \EndFor
    
    \State \textbf{Return} $\theta^{best}$ \Comment{Return best parameters found}
\end{algorithmic}
\end{algorithm}

\section{Experiment} \label{Experiment}


\subsection{Dataset Description}
In our experiment, we utilize two datasets: the publicly available WiGesture dataset \cite{CSI-BERT} and a novel dataset named WiFall, which is proposed along with this paper. Both datasets are collected using the ESP32-S3 device\textcolor{black}{, which operates in the 2.4 GHz Wi-Fi band-the default central frequency supported by the hardware} and share the same data format, with a sample rate of 100Hz, 1 antenna, and 52 subcarriers. For our experiment, we divide the data into 1-second samples, resulting in each sample having a length of 100 data points. 


\subsubsection{WiGesture Dataset} \label{WiGesture Dataset}
The WiGesture Dataset is a gesture recognition and people identification dataset. We use the dynamic part of the dataset that {\color{black} includes} 6 action id {\color{black} including left-right, forward-backward, and up-down motions, clockwise circling, clapping, and waving} and 8 people id. The dataset is collected by 8 college students in a meeting room. The ESP32-S3 serves as receiver and a home Wi-Fi router serves as transmitter. The six actions include left-right, forward-backward, and up-down motions, clockwise circling, clapping, and waving. In the gesture recognition task, we consider different people as distinct domains. Similarly, in the people identification task, we treat different actions as separate domains. 

\subsubsection{WiFall Dataset} \label{WiFall Dataset}
\begin{figure}[htbp]
\centering 
\includegraphics[width=0.4\textwidth]{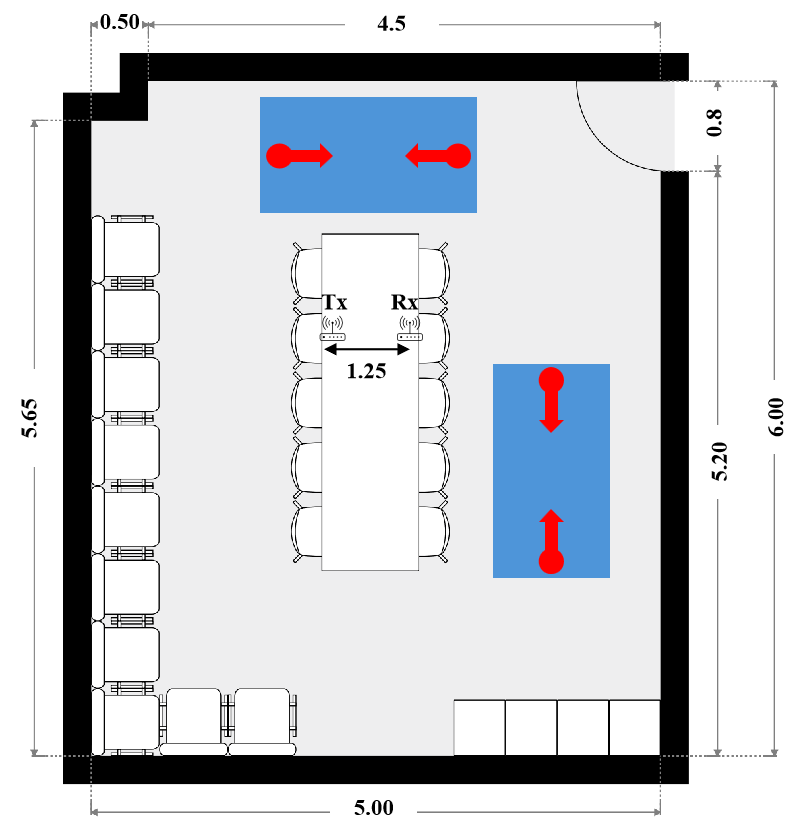}
\caption{Data Collection Environment of WiFall Dataset: The unit in the picture is meters.}
\label{env}
\end{figure}

\begin{figure*}
\centering 
\subfloat[Walk]{\includegraphics[width=0.2\textwidth]{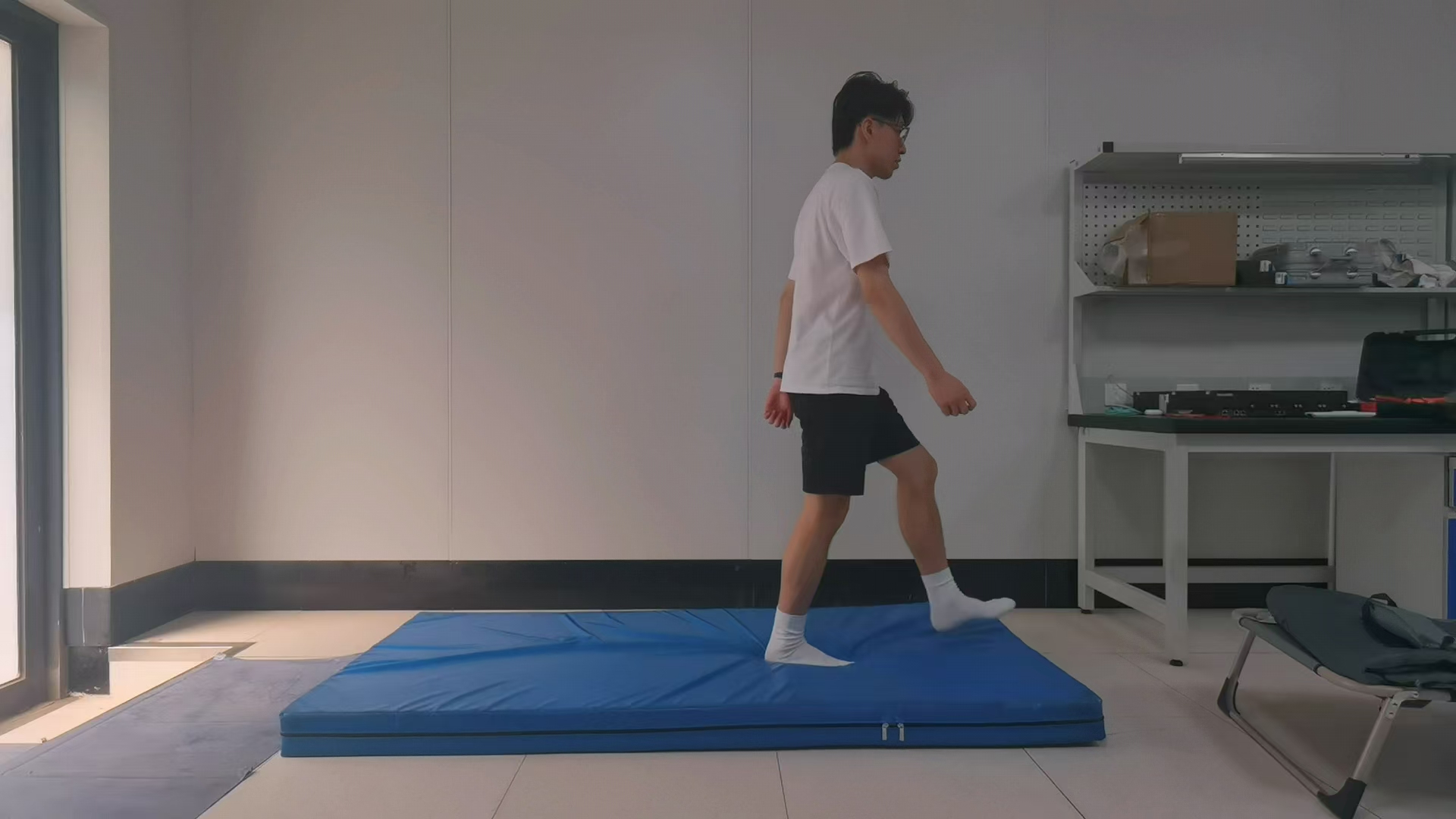}}
\subfloat[Jump]{\includegraphics[width=0.2\textwidth]{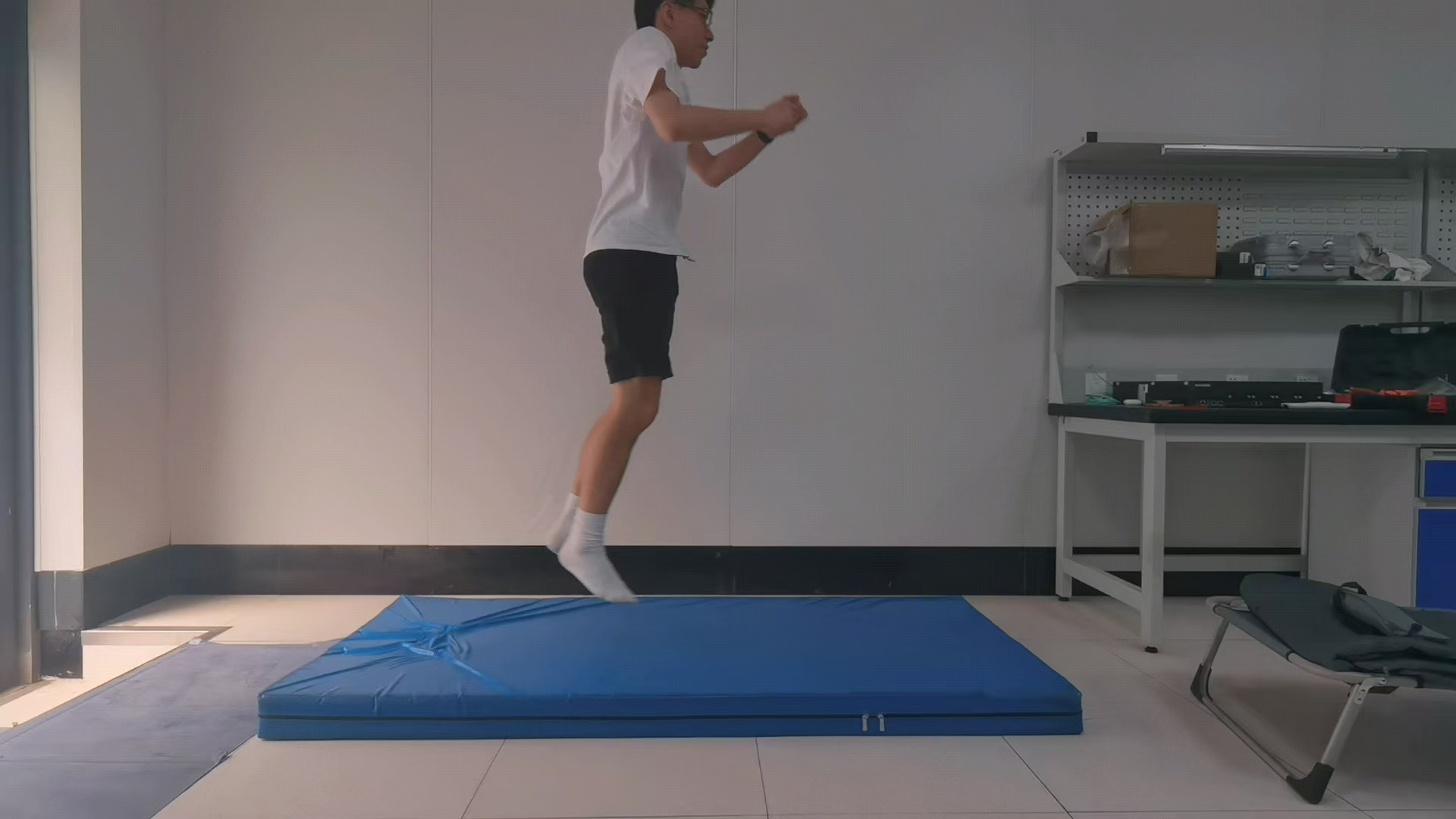}}
\subfloat[Sit]{\includegraphics[width=0.2\textwidth]{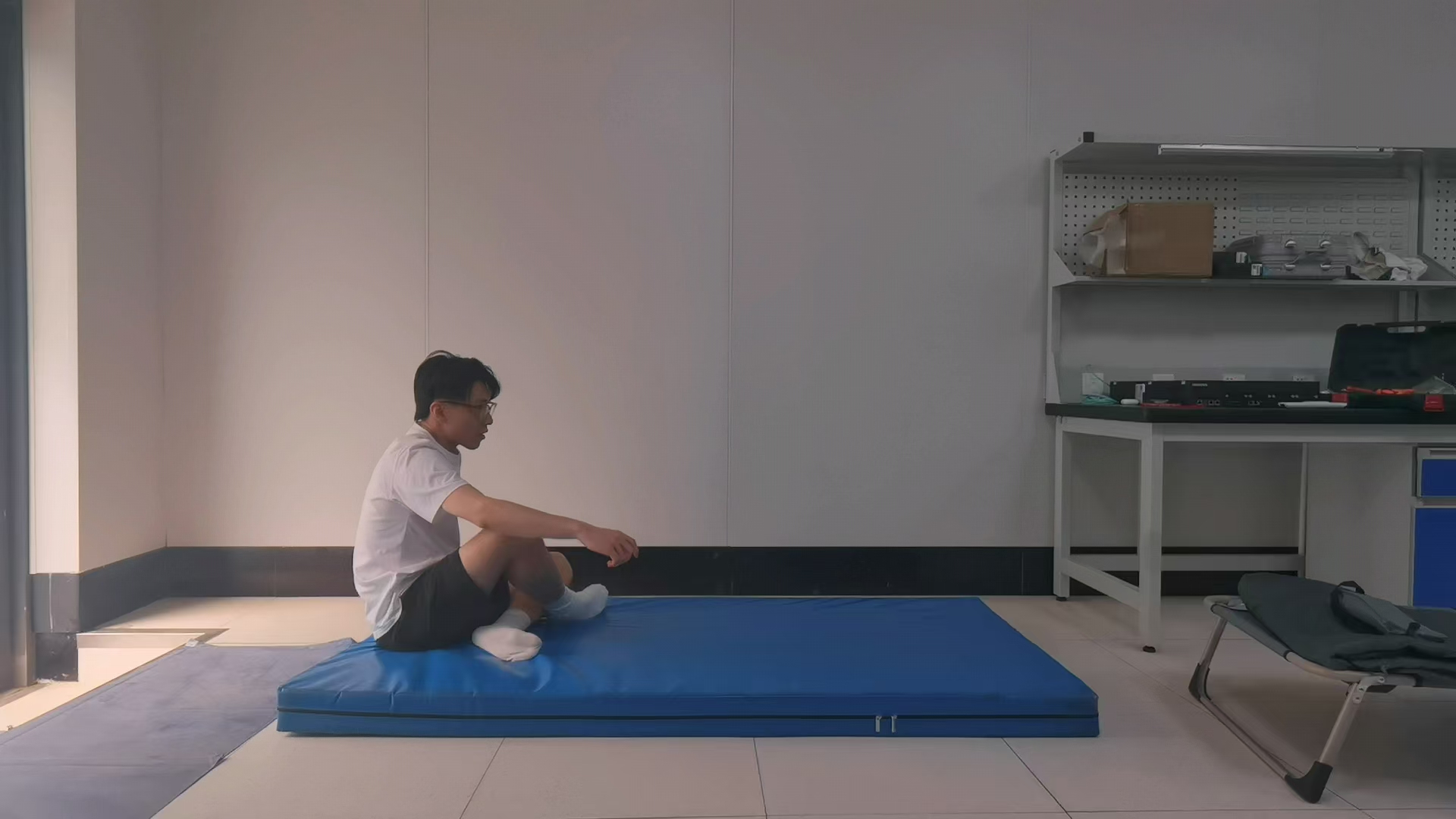}}
\subfloat[Stand Up]{\includegraphics[width=0.2\textwidth]{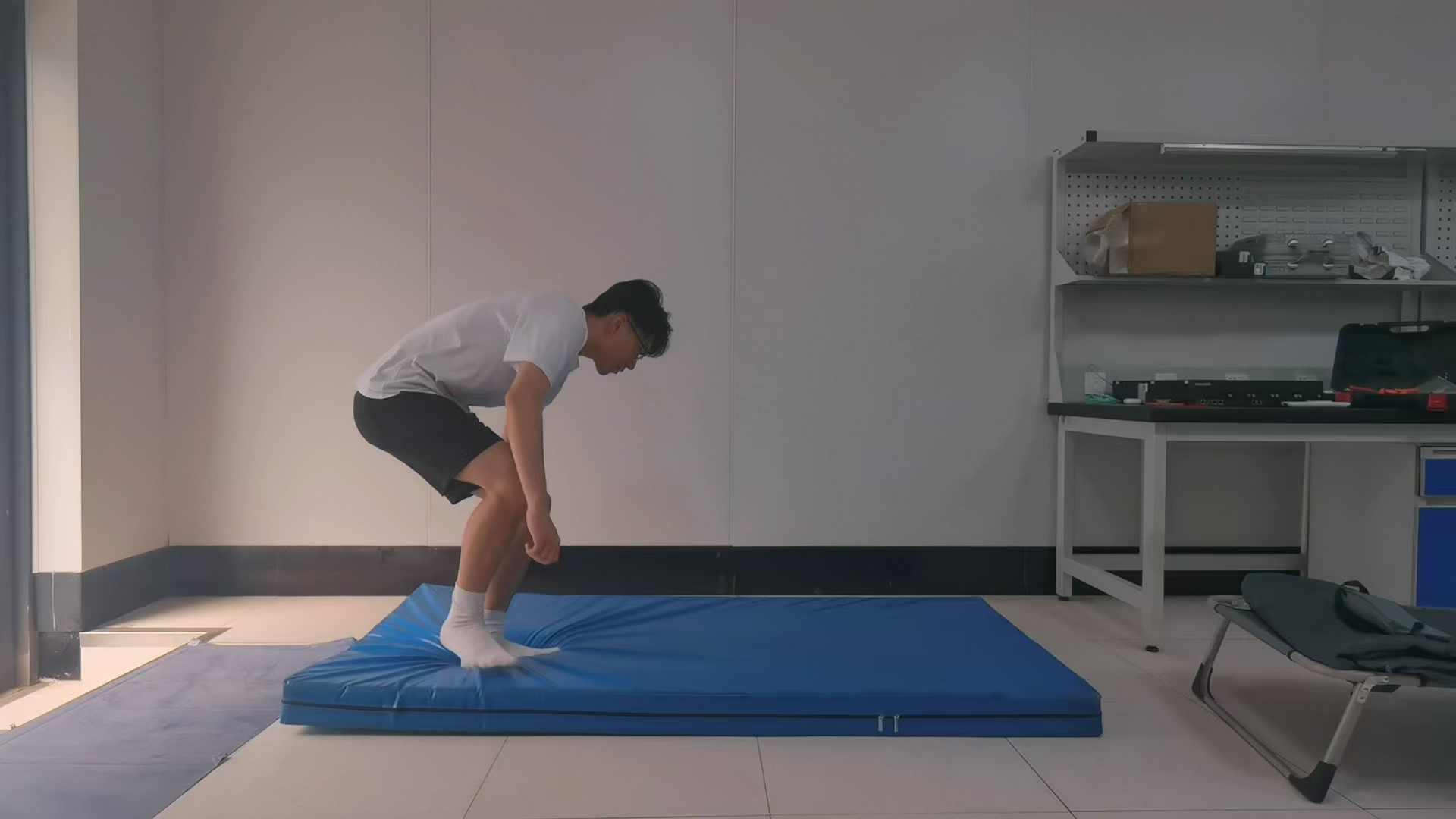}}
\subfloat[Fall]{\includegraphics[width=0.2\textwidth]{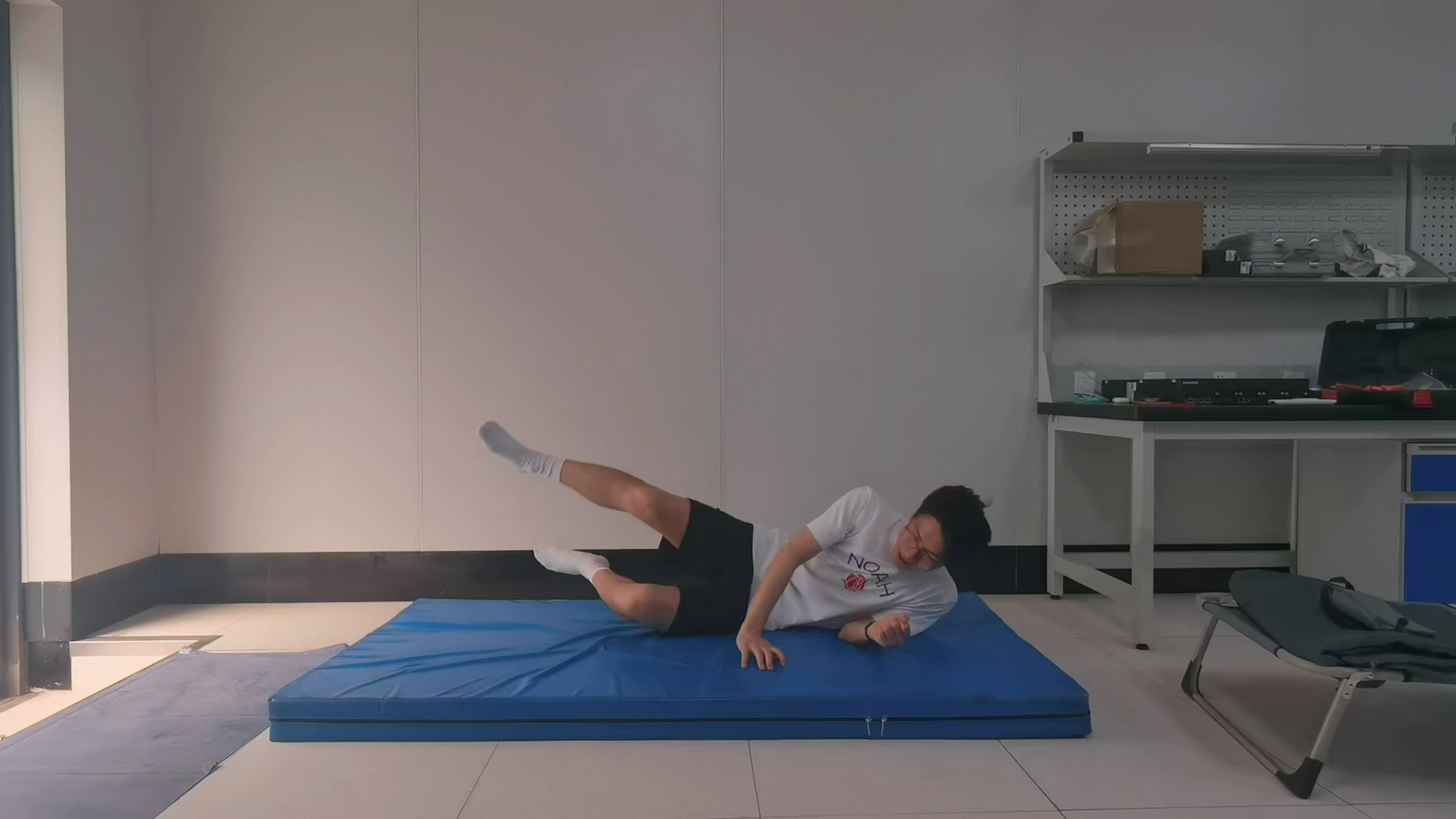}}
\caption{Actions of WiFall Dataset}
\label{WiFall}
\end{figure*}

The data collection scenario and device are the same as the WiGesture dataset, as shown in Fig. \ref{env}. In the dataset, we have a total of 10 volunteers with an average age of 23.17, height of 1.75m, and BMI of 23.76. {\color{black} The activities include 5 types: walking, jumping, sitting, standing up, and falling}, which also consists of many types like forward fall, left fall, right fall, seated fall, and backward fall, shown as Fig. \ref{WiFall}. Similar to the WiGesture Dataset, we consider different people as distinct domains. In our experiment, we use WiFall dataset to conduct action recognition tasks, where we identify 5 different actions, as well as fall detection tasks, where we only distinguish between fall and other actions.

\subsection{Experiment Setup}
Before conducting the experiment, we define the training set, support set, and testing set for each task in the n-shot scenarios as shown in Table \ref{scenario}. 

{\color{black} In the experiment, we split the dataset into segments of 1 second each. In the WiGesture dataset, each action performed by each individual lasts for 1 minute, resulting in 60 samples per action. There are 6 gestures and 8 participants in total. For the gesture recognition task, we use participants with IDs 1-7 as the training set, which includes 420 samples for each class, totaling 2,520 samples. The last 360 samples (60 samples for each class) from participant ID 0 are used as the target domain. In the people identification task, we use action IDs 1-5 for the training set, which consists of 300 samples per class, totaling 2,400 samples. The last 480 samples (60 samples for each class) from action ID 0 serve as the target domain.

In the WiFall dataset, each action performed by each individual lasts for 12 seconds, resulting in 12 samples, except for the action ``fall". For the ``fall" action, each participant repeats it 48 times, with each repetition lasting 1 second, corresponding to 1 sample. There are 5 actions, including ``fall", across 10 participants. For the fall detection task, we combine all non-fall actions into a single category and use participants with IDs 1-9 as the training set, which includes 432 samples for each class, totaling 864 samples. The last 96 samples (48 samples for each class) from participant ID 0 are used as the target domain. In the action recognition task, to maintain label balance between classes, we randomly drop $\frac{3}{4}$ of the samples from the ``fall" class. We then use participants with IDs 1-9 as the training set, which includes 108 samples per class, totaling 540 samples. The last 60 samples (12 samples for each class) from participant ID 0 are used as the target domain.
}

\begin{table*}
\caption{\color{black}{n-shot Scenario Description}}
\centering
\begin{adjustbox}{width=1.00\textwidth}
\begin{tabular}{|c||c|c|c|c|}
\hline
\textbf{Task} & \textbf{Dataset} & \textbf{Training Set} & \textbf{Support Set (Also Used as Valid Set)} & \textbf{Testing Set}\\
\hline 
\textbf{Gesture Recognition} & WiGesture & People ID 1-7 & $n$ samples for each gesture in People ID 0 & samples excluding support set in People ID 0\\
\hline
\textbf{People Identification} & WiGesture  & Action ID 1-5 & $n$ samples for each person in Action ID 0 & samples excluding support set in Action ID 0\\
\hline
\textbf{Fall Detection} &  WiFall & People ID 1-9 & $n$ samples for each action in People ID 0 & samples excluding support set in People ID 0\\
\hline
\textbf{Action Recognition} & WiFall & People ID 1-9 & $n$ samples for each action in People ID 0 & samples excluding support set in People ID 0\\
\hline \hline 
\textbf{Task} & \textbf{Class Amount} & \textbf{Training Set Size} & \textbf{Support Set Size} & \textbf{Testing Set Size}\\
\hline 
\textbf{Gesture Recognition} & 6 & $6 \times 420 = 2,520$ & $6 \times n$ & $6 \times (60-n)$ \\
\hline
\textbf{People Identification} & 8 & $8 \times 300 = 2,400 $ & $8 \times n$ & $8 \times (60-n)$ \\
\hline
\textbf{Fall Detection} & 2 & $2 \times 432 = 864$ & $2 \times n$ & $2 \times (48-n)$ \\
\hline
\textbf{Action Recognition} & 5 & $5 \times 108 = 540$ & $5 \times n$ & $5 \times (12-n)$ \\
\hline
\end{tabular}
\end{adjustbox}
\label{scenario}
\end{table*}

\begin{table}[htbp]
\caption{Model Configurations: This table provides a detailed overview of our KNN-MMD in the following experiments.}
    \centering
        \begin{tabular}{|c|c|}
        \hline
        \textbf{Configuration} & \textbf{Our Setting} \\
        \hline 
        CSI Sample Length     & 100   \\
        \hline 
        CSI Subcarrier Number     & 52   \\
        \hline 
        Batch Size   & 256  \\
        \hline
        Input Dimension   & (256,1,100,52)   \\
        \hline 
        Hidden Dimension of UMAP \cite{UMAP}   & 128    \\
        \hline 
        Hidden Dimension of ResNet \cite{ResNet}  & 64   \\
        \hline 
        Neighbor Number in KNN \cite{KNN} : $k$   & 1   \\
        \hline 
        Selected Proportion in \textcolor{black}{Help Set} : $p\%$  & 50\%    \\
        \hline 
        Optimizer    & Adam \\
        \hline 
        Learning Rate    & 0.0005  \\
        \hline 
        Hyper-parameters in Early Stop (Algorithm \ref{alg:early_stop})   & \multirow{2}{*}{[200,350,30,1.2,0.8]}  \\
        $[e^{min},e^{max},e^{threshold},\alpha,\beta]$ & \\
        \hline 
        \multirow{2}{*}{Kernel List $K$}    & Gaussian Kernels  \\
        & ($\sigma=0.5,1$) \\
        \hline 
        Total Number of Parameters  & 11.21 million   \\
        \hline 
        \end{tabular}
\label{configuration}
\end{table}

The implementation details of our model are presented in Table \ref{configuration}. For our experiments, we utilized an Intel(R) Xeon(R) Silver 4210R CPU @ 2.40GHz and a NVIDIA RTX 3090 GPU as the hardware devices. During training, we observed that our model consumed approximately 2,500 MB of GPU memory.

\color{black}
To evaluate the efficiency of our method, we compare it with several representative cross-domain methods in the fields of Wi-Fi sensing and machine learning. For fairness, we have made every effort to replicate other works according to the descriptions provided in their respective papers. For any missing settings in these papers, we adopt the same settings as our method.
In the field of Wi-Fi sensing, most works are based on the feature learning (learning-based) paradigm. We choose the four works mentioned before for comparison: Ding et al. \cite{ding2021wi}, AutoFi \cite{AutoFi}, Yang et al. \cite{yang2019learning}, and CrossFi \cite{CSi-Net}, which represent different powerful feature extractors and similarity calculation techniques. Even some works combined with DAL methods, such as Yang et al. \cite{yang2019learning} and CrossFi \cite{CSi-Net}, which utilize the MK-MMD loss, do not rely solely on it.
To further illustrate the performance of the DAL methods, we have chosen several popular methods in machine learning, including adversarial learning-based DANN \cite{DANN} and ADDA \cite{ADDA}, as well as feature-space alignment methods such as MK-MMD \cite{MK-MMD} and GFK \cite{GFK}. Since these methods do not depend on specific network structures, we use the same ResNet architecture as in this paper. Additionally, we employ the same kernel functions in MK-MMD as in our KNN-MMD method.
Furthermore, we also compare the methods of Tian et al. \cite{KNN-MMD} and EEG \cite{EEG}, which also discuss local alignment methods, with detailed descriptions provided in Section \ref{Domain Adaptation Methods}. All these DAL methods fall under the zero-shot category, meaning they only require a labeled training set from the source domain and an unlabeled testing set, without the necessity for a labeled support set.
\color{black}

\subsection{Experiment Result} \label{Experiment Result}
First, we present the experimental results in the one-shot and zero-shot scenarios as shown in Table \ref{tab:result}. Although ResNet18 \cite{ResNet} is not specifically designed for cross-domain scenarios, it is commonly used as a feature extractor in many related methods. Therefore, we include the results of ResNet in the in-domain scenario and zero-shot scenario as benchmarks. It can be seen that our KNN-MMD method shows excellent performance in each task, particularly outperforming all other methods in gesture recognition and action recognition tasks.

\begin{figure*} [t!]
\centering 
\subfloat[Gesture Recognition]{\includegraphics[width=0.45\textwidth]{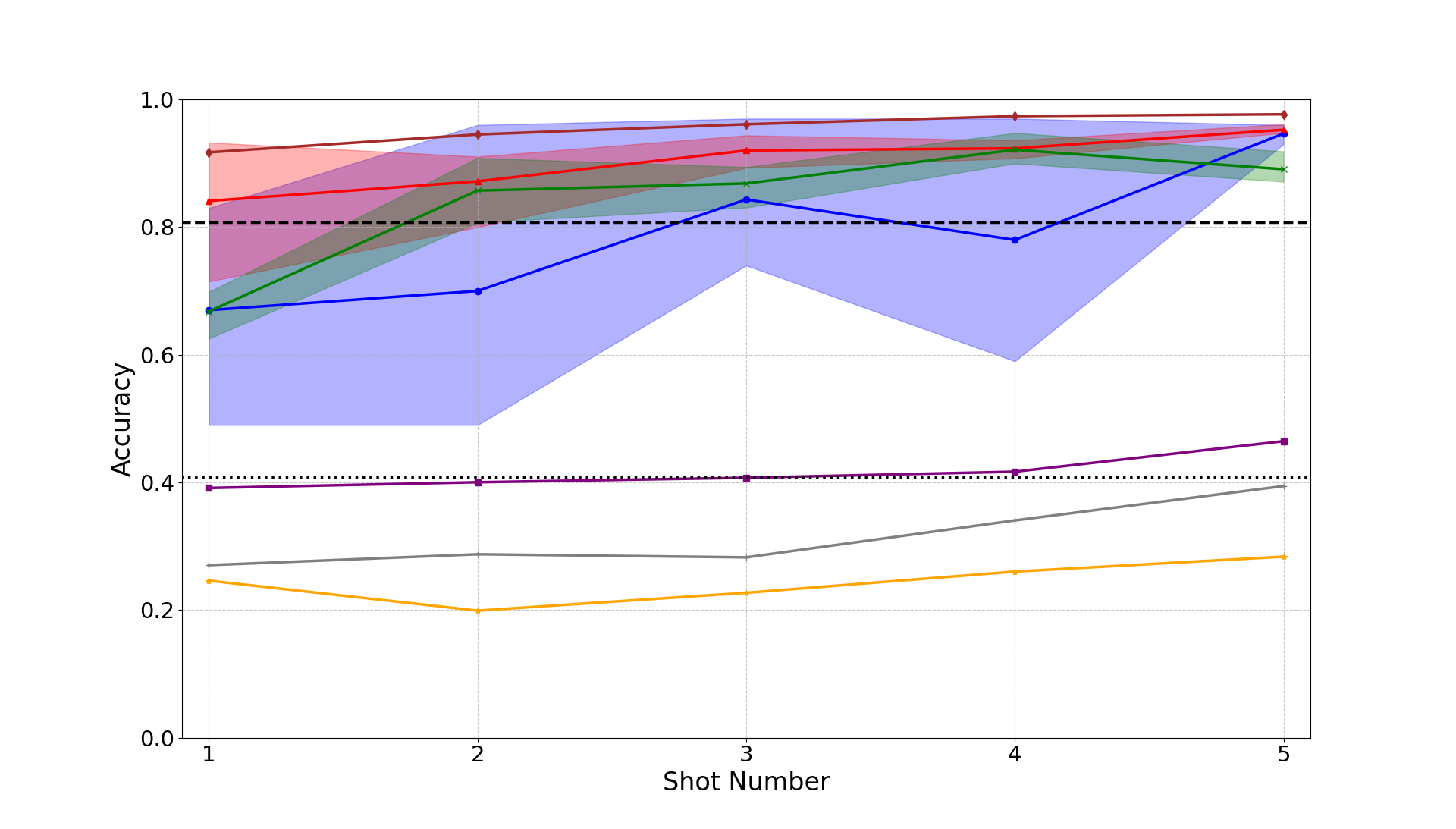}}
\subfloat[People Identification]{\includegraphics[width=0.45\textwidth]{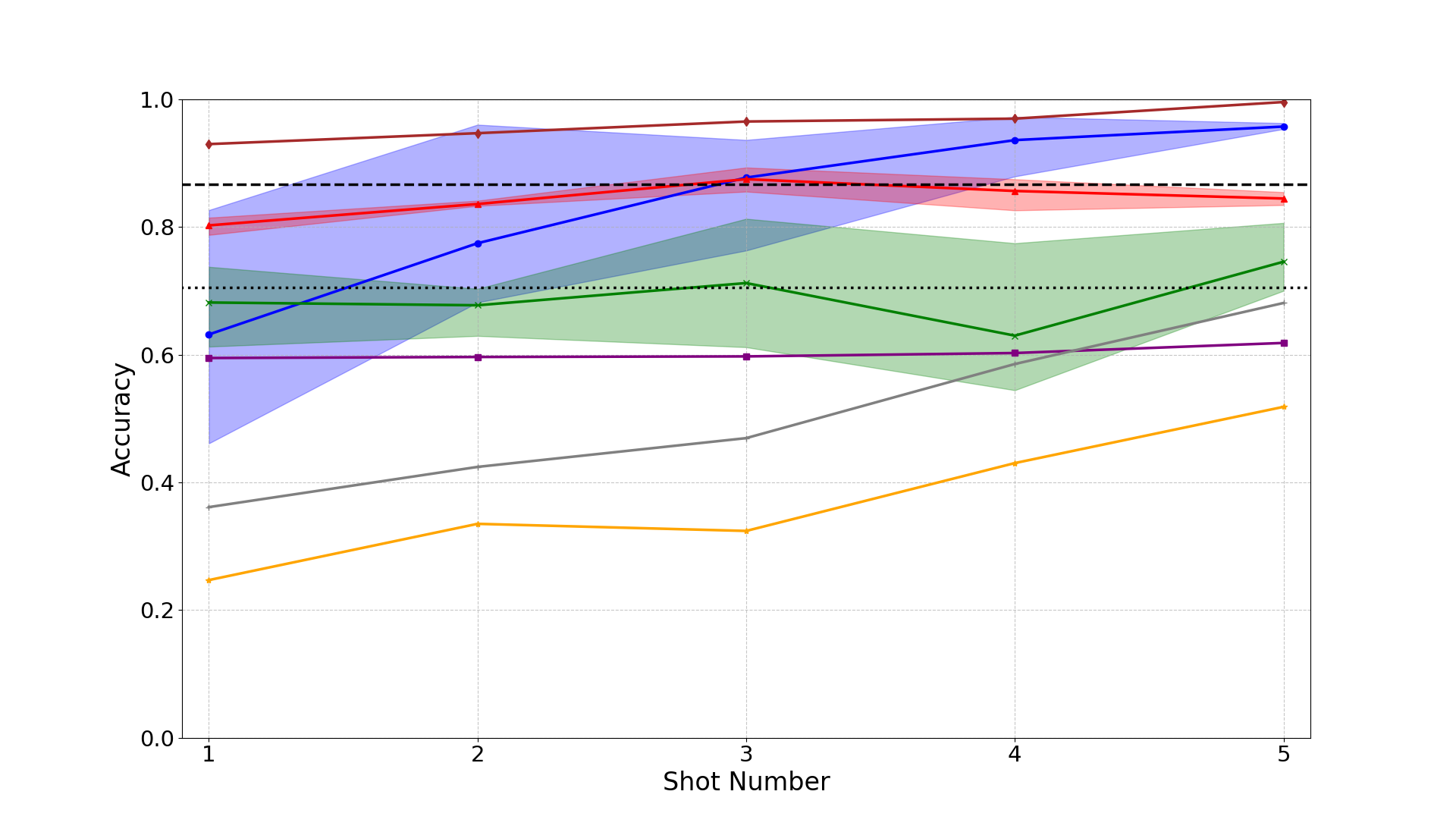}}
\\
\subfloat[Fall Detection]{\includegraphics[width=0.45\textwidth]{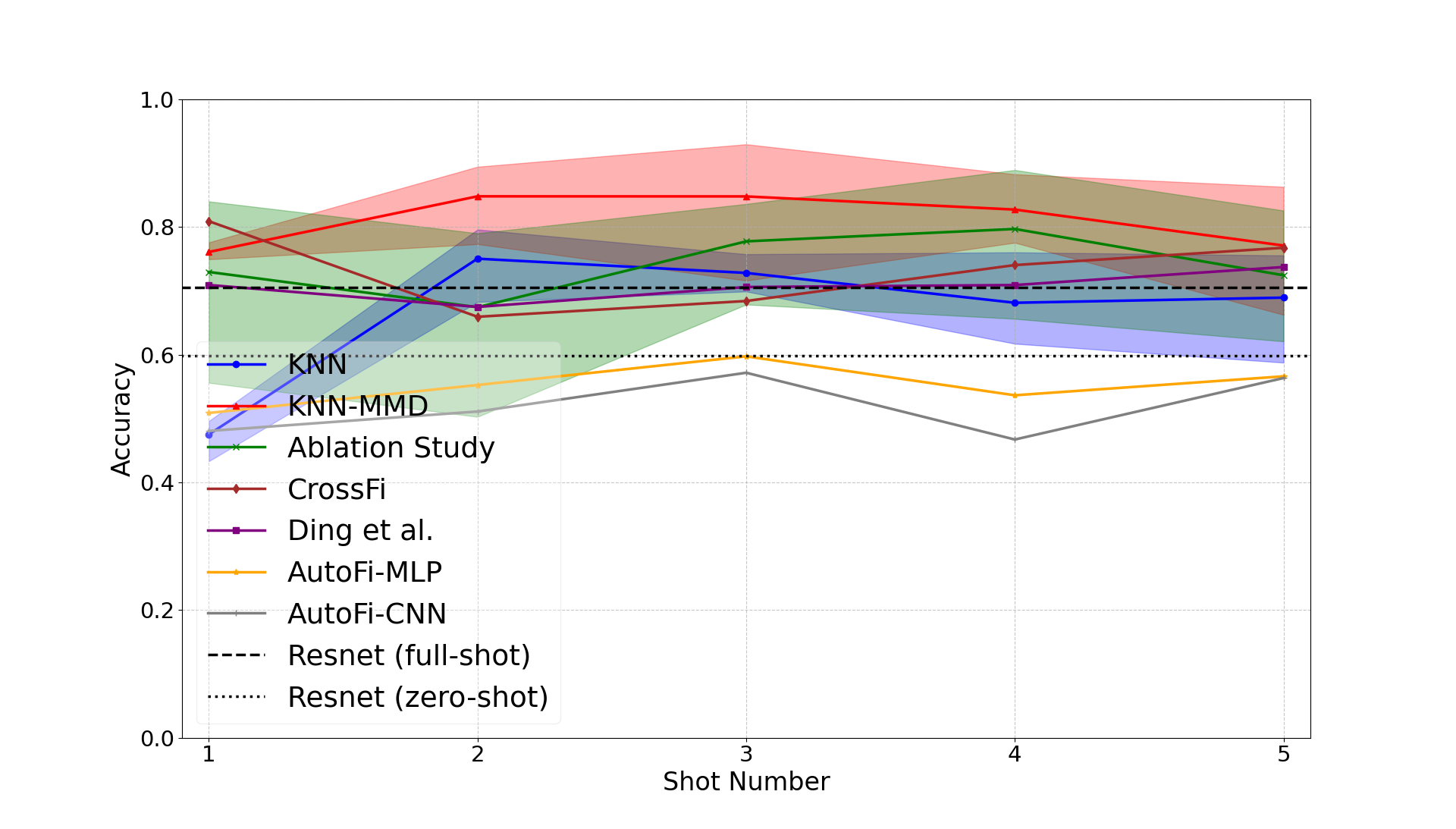}}
\subfloat[Action Recognition]{\includegraphics[width=0.45\textwidth]{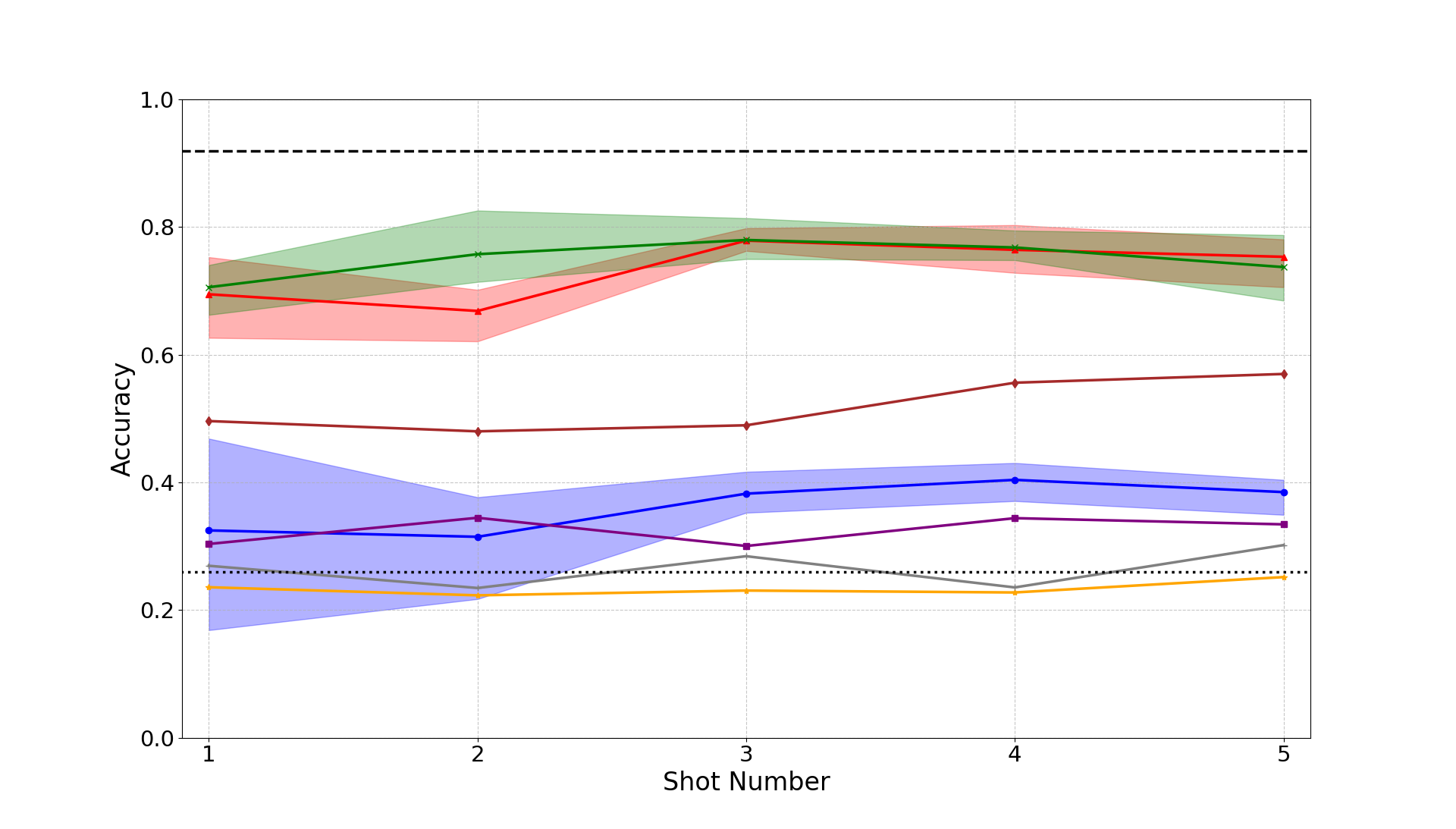}}
\caption{ \color{black}{Few-shot Experiment: The figures illustrate the influence of the number of shots on testing accuracy. The shaded area represents the range of accuracy observed across multiple experiments.}}
\label{few-shot result}
\end{figure*}

Then, as shown in Fig. \ref{few-shot result}, we illustrate the impact of the number of shots on classification accuracy. To assess the stability of our model, we also present shaded curves for our method and KNN by conducting five experiments. As mentioned earlier, it is challenging for other FSL methods to determine when the model is converged. Therefore, we directly use the best testing accuracy achieved by a particular FSL method during training as its accuracy. Even with this potentially unfair comparison mechanism, our KNN-MMD approach outperforms most conventional methods in all scenarios, which proves the efficiency of our local distribution approach. What's more, from the figure, it can be also observed that our model exhibits smaller fluctuations compared to vanilla KNN. This can be attributed to our preliminary classification mechanism. By using the confidence level to select the help set, our method's performance is less sensitive to the quality of the support set.


\begin{table*}
\caption{Experimental Results: Comparing the Best Performance of Each Method. The bold value represents the consistently superior result within each scenario, maintained across subsequent tables.}
    \centering
    \begin{adjustbox}{width=1.00\textwidth}
        \begin{tabular}{|c||c||c|c|c|c||c|}
        \hline
        \textbf{Method} & \textbf{Scenario} & \textbf{Gesture Recognition} & \textbf{People Identification} & \textbf{Fall Detection} & \textbf{Action Recognition} & \textbf{Average Accuracy} \\
        \hline 
        \multirow{2}{*}{\textbf{ResNet18\cite{ResNet}}} & in-domain & 80.75\% & 86.75\% & 91.88\% & 70.50\%  & 82.47\%  \\
        \cline{2-7}
         & zero-shot & 40.84\% & 70.50\% & 59.86\% & 26.00\% &  49.30\%\\
        \hline \hline
        \textbf{Siamese\cite{siamese}} & one-shot & 70.40\% & 82.87\% & 60.62\% & 38.95\%  & 63.21\% \\
        \hline
        \textbf{AutoFi (MLP-based)\cite{AutoFi}} & one-shot & 24.62\% & 24.71\% & 50.88\% & 23.59\%  &  30.95\% \\
        \hline
        \textbf{AutoFi (CNN-based)\cite{AutoFi}} & one-shot &  27.05\% & 36.14\%  & 48.05\% & 26.95\%  & 34.55\% \\
        \hline
        \textbf{Yang et al.\cite{SN-MMD}} & one-shot & 67.21\% & 74.22\% & 59.75\% & 48.52\%  & 62.43\% \\
        \hline
        \textbf{Ding et al.\cite{CNN-based}} & one-shot & 39.14\% & 70.94\% & 61.56\% & 30.37\%  & 50.50\% \\
        \hline
        \textbf{CrossFi\cite{CSi-Net}} & one-shot & 91.72\% & \textbf{93.01\%} & 80.93\% & 49.62\% & 78.82\% \\
        \hline
        \textbf{KNN \cite{KNN}} & one-shot & 83.02\% & 82.67\% & 49.63\% & 46.87\% & 65.55\% \\
        \hline
        \textbf{KNN-MMD (Ours)} & one-shot & \textbf{93.26\%} & 81.84\% & 77.62\% & \textbf{75.30\%}  & \textbf{82.01\%} \\
        \hline 
        \textbf{Ablation Study} & one-shot & 69.87\% & 73.78\% & \textbf{84.03\%} & 74.06\% & 75.44\% \\
        \hline \hline 
        \textbf{MK-MMD\cite{MK-MMD}} & zero-shot & 40.36\% & 66.47\% & 72.26\% & 43.72\% & 55.70\% \\
        \hline
        \textbf{DANN\cite{DANN}} & zero-shot & 41.41\% & 67.18\% & 74.06\% & 35.99\%  & 54.66\% \\
        \hline
        \textbf{ADDA\cite{ADDA}} & zero-shot & 42.71\% & 65.43\% & 62.81\% & 36.08\% & 51.76\% \\
        \hline
        \textbf{GFK+KNN\cite{GFK}} & zero-shot & 30.79\% & 51.50\% & 53.72\% & 34.17\% & 42.55\% \\
        \hline 
       \textbf{CrossFi\cite{CSi-Net}} & zero-shot & 64.81\% & \textbf{72.79\%} & \textbf{74.38\%} & 40.46\%  & \textbf{63.11\%} \\
        \hline  
        \textbf{Tian et al.\cite{KNN-MMD}} & zero-shot  & \textbf{68.13\%} & 55.86\% &  61.72\% & 42.10\% & 56.95\% \\
        \hline
        \textbf{EEG\cite{EEG}} & zero-shot & 59.75\% & 64.63\% & 69.53\% & 42.15\% & 59.02\% \\
        \hline 
        \end{tabular}
    \end{adjustbox}
\label{tab:result}
\end{table*}

\subsection{Comparison with Traditional DA Methods} \label{Comparison with Traditional DA Methods}
In this section, we use the gesture recognition task to illustrate the shortcomings of traditional DA methods. For each type of method, we choose a specific model and compare it with our KNN-MMD.

\subsubsection{Metric-based Method}

For metric-based methods like KNN \cite{KNN}, the biggest problem is that they are heavily influenced by the quality of the support set. 
As shown in Table \ref{tab:KNN}, we reduced the data dimension in the target domain to $d$ using UMAP and tested the KNN performance in an $n$-shot scenario with $k$ nearest neighbors. Each scenario was tested 3 times, and the table shows the accuracy range. It can be seen that the accuracy fluctuates greatly, which may pose challenges for the practical application of this method. In Fig. \ref{few-shot result}, it can be seen that our KNN-MMD method does not exhibit such large fluctuations. Moreover, in Table \ref{tab:KNN}, it's noticeable that our KNN-MMD is not very sensitive to the hyper-parameters.


\begin{table*}
\caption{Performance of KNN \cite{KNN} and Our KNN-MMD:  $n$ denotes the number of shots, $k$ denotes the number of neighbors in KNN, and $d$ denotes the data dimension after reduction using UMAP \cite{UMAP}.}
    \centering
        \begin{tabular}{|c||c|c|c||c|c|c|}
        \hline
        & \multicolumn{3}{c||}{KNN \cite{KNN}} & \multicolumn{3}{c|}{KNN-MMD}\\
        \cline{2-7}
         & \textbf{d=32} & \textbf{d=64} & \textbf{d=128} & \textbf{d=32} & \textbf{d=64} & \textbf{d=128}\\
        \hline 
        \textbf{n=1, k=1} & 49\%-83\% & 49\%-69\% & 51\%-83\%  & 85\%-95\% & 83\%-93\% & 72\%-93\% \\
        \hline
        \textbf{n=2, k=1} &  72\%-92\% & 79\%-89\% & 65\%-96\% & 79\%-94\% & 87\%-93\% & 80\%-91\%\\
        \hline 
        \textbf{n=2, k=2} &  65\%-76\% & 58\%-82\% & 49\%-83\% & 88\%-95\% & 84\%-92\% & 88\%-91\% \\
        \hline
        \textbf{n=3, k=1} &  78\%-94\% & 74\%-93\% & 91\%-97\% & 88\%-95\% & 87\%-95\% & 89\%-92\% \\
        \hline 
        \textbf{n=3, k=2} &  74\%-94\% & 68\%-97\% & 77\%-82\% & 87\%-93\% & 87\%-91\%  & 84\%-92\% \\
        \hline 
        \textbf{n=3, k=3} &  64\%-93\% & 68\%-94\% & 74\%-91\% & 91\%-96\% & 86\%-90\% & 90\%-94\% \\
        \hline
        \textbf{n=4, k=1} &  83\%-97\% & 77\%-97\% & 78\%-97\% & 92\%-96\% & 88\%-92\% & 90\%-93\% \\
        \hline 
        \textbf{n=4, k=2} &  91\%-96\% & 92\%-97\% & 80\%-95\% & 94\%-98\% & 91\%-94\% & 91\%-96\% \\
        \hline 
        \textbf{n=4, k=3} &  77\%-97\% & 73\%-97\% & 82\%-92\% & 85\%-93\% & 89\%-93\% & 90\%-94\% \\
        \hline 
        \textbf{n=4, k=4} &  80\%-95\% & 59\%-88\% & 59\%-97\% & 87\%-94\% & 90\%-94\% &  88\%-95\% \\
        \hline 
        \end{tabular}
\label{tab:KNN}
\end{table*}

\subsubsection{Learning-based Method}

\begin{figure}[htbp]
\centering 
\subfloat[Siamese Network \cite{siamese}]{\label{siamese_train}\includegraphics[width=0.4\textwidth]{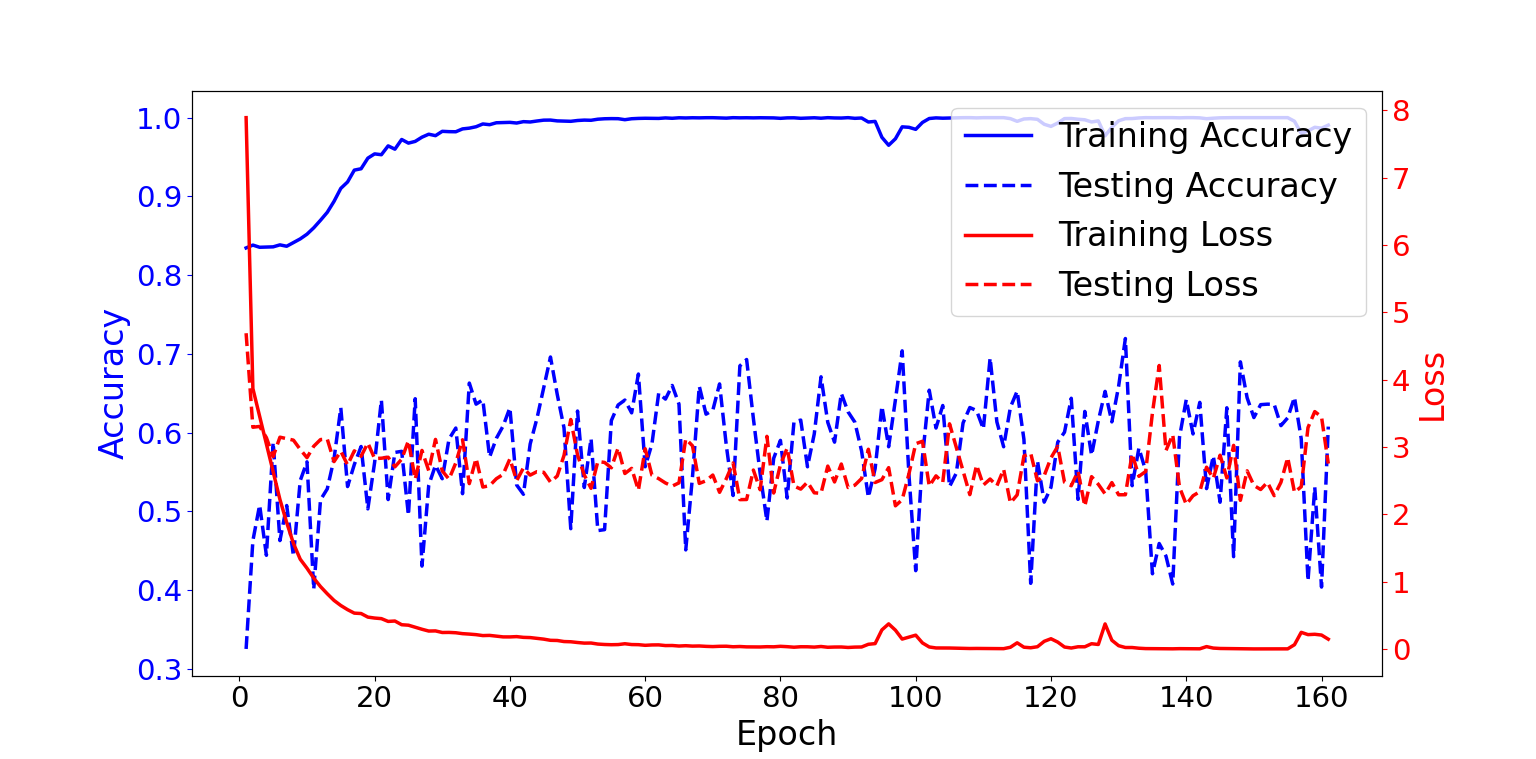}}\\
\subfloat[MK-MMD \cite{MK-MMD}]{\label{mkmmd_train}\includegraphics[width=0.4\textwidth]{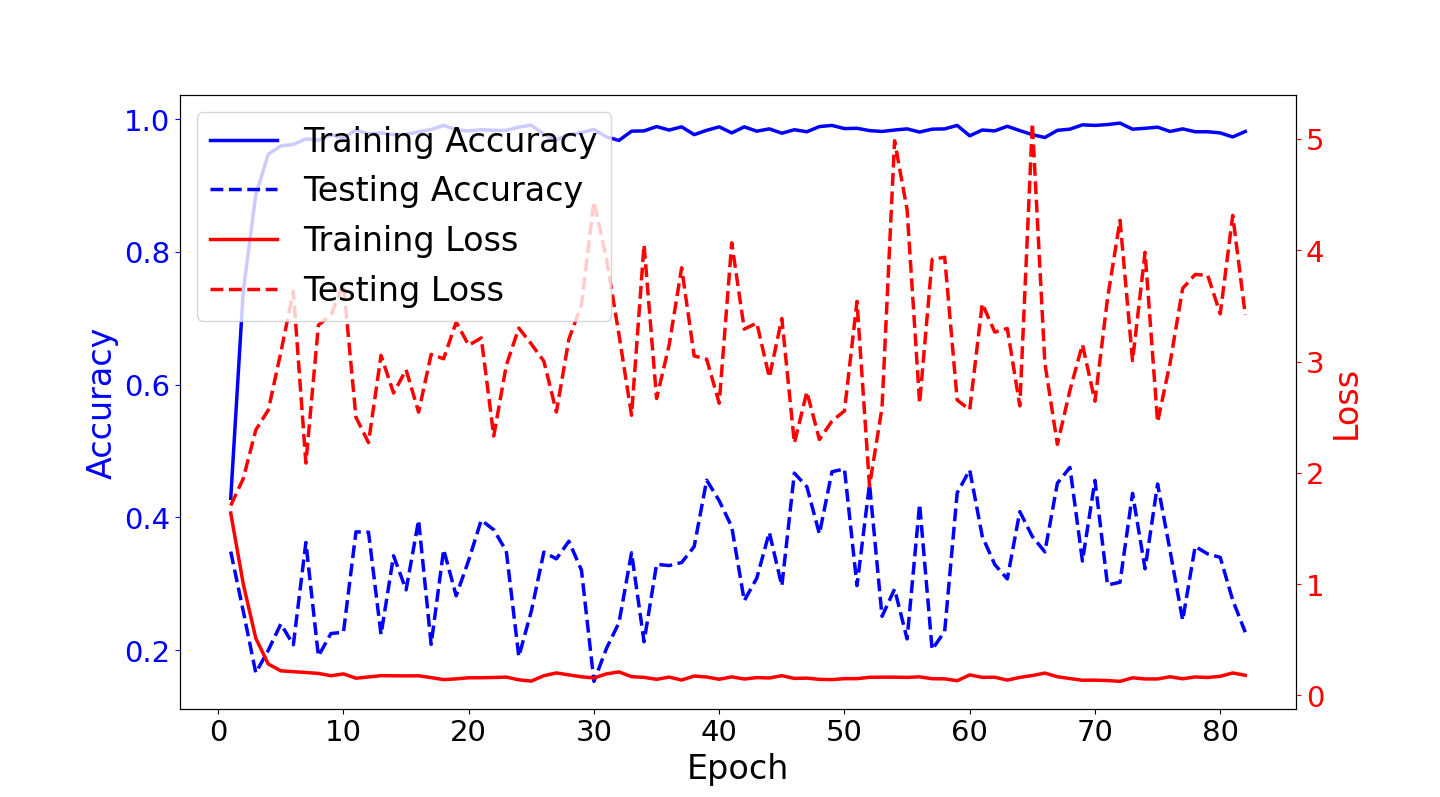}}
\\
\subfloat[KNN-MMD (shot number=1)]{\label{knnmmd_train}\includegraphics[width=0.4\textwidth]{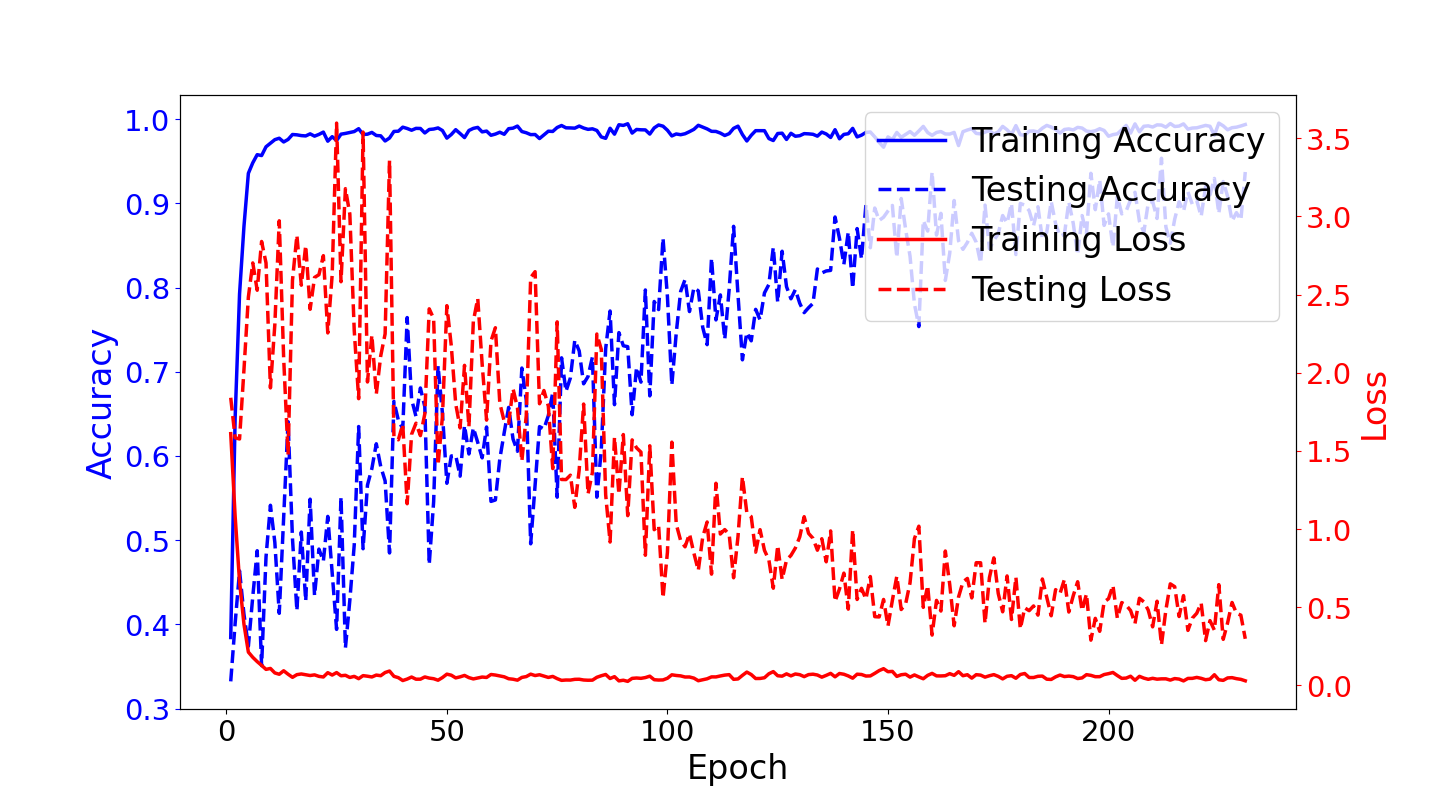}}
\caption{Training Process of Different Methods} 
\label{fig:train_process}
\end{figure}

For learning-based methods, the biggest problem is that we do not know when to stop the training process. Fig. \ref{siamese_train} illustrates the training process of a Siamese network \cite{siamese}, a traditional one-shot learning model. It can be seen that the performance change is relatively stable on the training set, but not on the testing set. However, during training, we do not have direct access to the performance on the testing set. This makes it very possible to stop the model training at an epoch that has worse performance on the testing set. Our KNN-MMD method efficiently solves this problem by using early stop according to the accuracy in the support set. In traditional learning-based methods, the support set also participates in training, so our early stop method cannot be used in them. Furthermore, as shown in Fig. \ref{knnmmd_train}, the testing accuracy of our KNN-MMD also increases stably with the help of our local alignment mechanism. 

\subsubsection{DAL Method}

\begin{figure*}[t!]
\centering 
\subfloat[ResNet \cite{ResNet} \\ (no alignment)]{\includegraphics[width=0.25\textwidth]{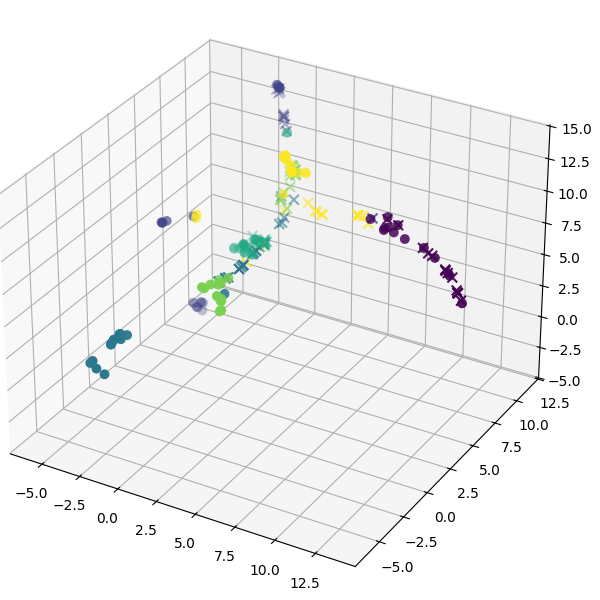}}
\subfloat[MK-MMD \cite{MK-MMD} \\ (global alignment)]{\includegraphics[width=0.25\textwidth]{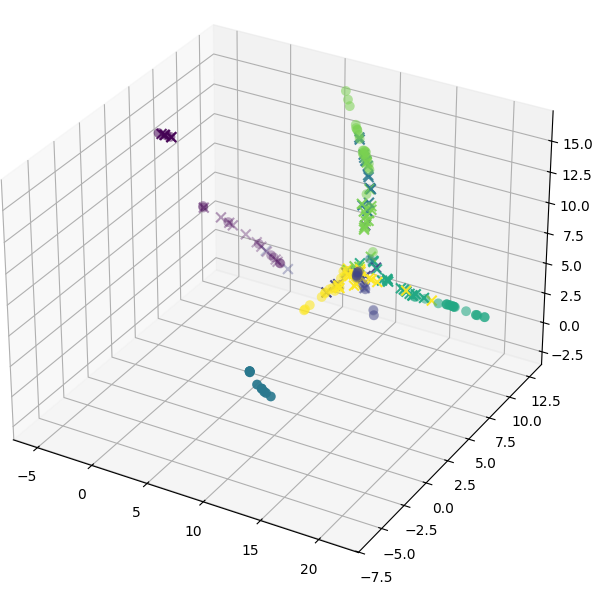}} 
\subfloat[KNN-MMD \\  (shot number=1)]{\label{UMAP-KNN-MMD}\includegraphics[width=0.25\textwidth]{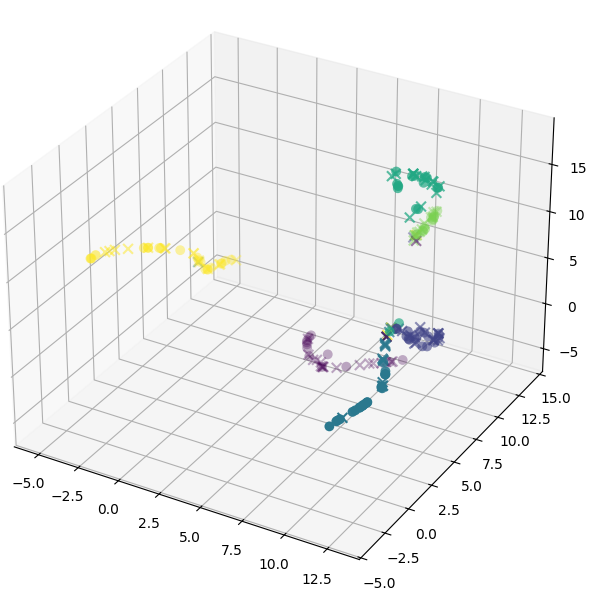}}
\subfloat[Theoretical Upper Bound \\  (local alignment)\label{Upper}]{\includegraphics[width=0.25\textwidth]{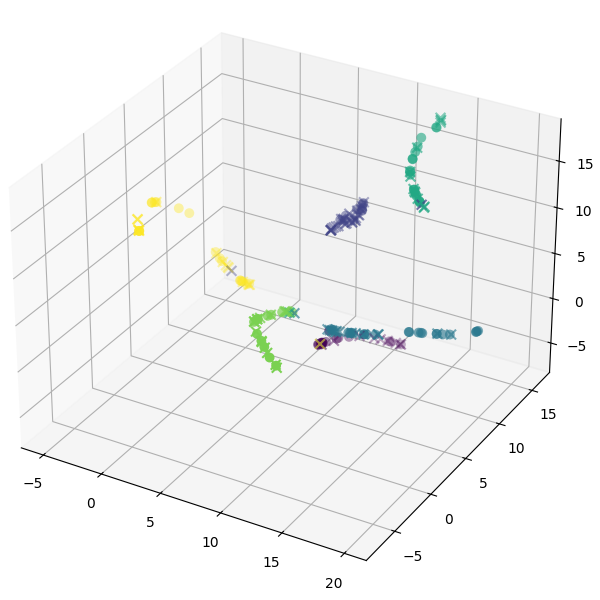}}
\caption{Data Dimension Reduction Results of Embedding Results from Different Models: Different colors represent different categories. The circles represent samples from the source domain, and the crosses represent samples from the target domain.}
\label{fig:umap}
\end{figure*}

As mentioned before, the global alignment operation of traditional DAL methods cannot guarantee the model performance. In Fig. \ref{fig:umap}, we illustrate the UMAP result of different methods, including: ResNet (no alignment), MK-MMD (global alignment), KNN-MMD (shot number=1), and Theoretical Upper Bound (local alignment).


It can be seen that the performance of the global alignment method MK-MMD does not differ much from ResNet, which even has no alignment operation. Even though they can split samples in the source domain (training set) well, achieving an accuracy over 99\%, they cluster samples in the target domain (testing set) chaotically. In contrast, our KNN-MMD still performs well in the target domain. As shown in Fig. \ref{UMAP-KNN-MMD}, it successfully aligns samples from the source domain and target domain within each category. Furthermore, in Fig. \ref{Upper}, we show the upper bound performance of KNN-MMD, where we provide the ground truth label of the help set instead of pseudo-labels. We notice that the cluster performance and accuracy of KNN-MMD and the upper bound do not have a significant difference, which proves the efficiency of our help set construction method. Finally, even though our early stop method can be used in most DAL methods, as the support set does not directly participate in training, limited by the model principle, even if we make it stop at the point with the best testing accuracy, it still has a poor performance, as shown in \textcolor{black}{Fig. \ref{mkmmd_train}}.

\subsection{Ablation Study}
In our KNN-MMD method, the MK-MMD seems unnecessary because we can simply pre-train the model on the training set and fine-tune it on the help set as 
many DA methods do. However, when choosing the top $p\%$ samples in the target domain, we cannot guarantee that the amounts of different category samples will be balanced. This imbalance may cause the long-tail problem during fine-tuning and significantly impact the network's performance. However, MMD is not affected by this issue. Furthermore, in extreme situations, some categories may be missing in the help set, which can be catastrophic for training the network. On the other hand, our method, which combines local MMD and global MMD, can mitigate this problem to some extent. We compare the performance of our method with that of fine-tuning a ResNet18 using help set in each task. As shown in Fig. \ref{few-shot result}, our MMD-based method shows better performance than the fine-tuning-based method. And in most tasks, our method also demonstrates better stability.


\color{black}
\subsection{Expanded Experiment}

\subsubsection{Evaluation on WiDar3.0}

To further illustrate the efficiency of our KNN-MMD, we conduct cross-domain experiments using the WiDar3.0 dataset \cite{zhang2021widar3}, which includes both cross-people and cross-room gesture recognition scenarios. In WiDar3.0, the Body-coordinate Velocity Profile (BVP) extracted from CSI is provided, with the dimension of $t \times 20 \times 20$, where $t$ represents the time length. To standardize the data dimensions, we utilize only the first 28 frames of each sample and filter out those samples that are shorter than 28 frames. 
For our experiment, we select the subsets as follows: \textbf{(i) Cross-People Scenario:} For the cross-person gesture task, we utilize data collected on November 9, 2018, which includes six gestures: Push\&Pull, sweep, clap, slide, draw, zigzag, and draw-N. We designate User 1 as the source domain and User 2 as the target domain. Both the source and target domains contain 3,000 samples. \textbf{(ii) Cross-Room Scenario:} For the cross-room gesture task, we utilize data collected on November 21 and November 27, 2018, from User 2, which includes six gestures: slide, draw-O, draw zigzag, draw-N, draw-triangle, and draw-rectangle. We designate Room 1 as the source domain and Room 2 as the target domain. The source domain contains 3,000 samples, while the target domain has 750 samples.

\begin{figure}[t!]
\centering 
\subfloat[Cross-People Scenario]{\includegraphics[width=0.4\textwidth]{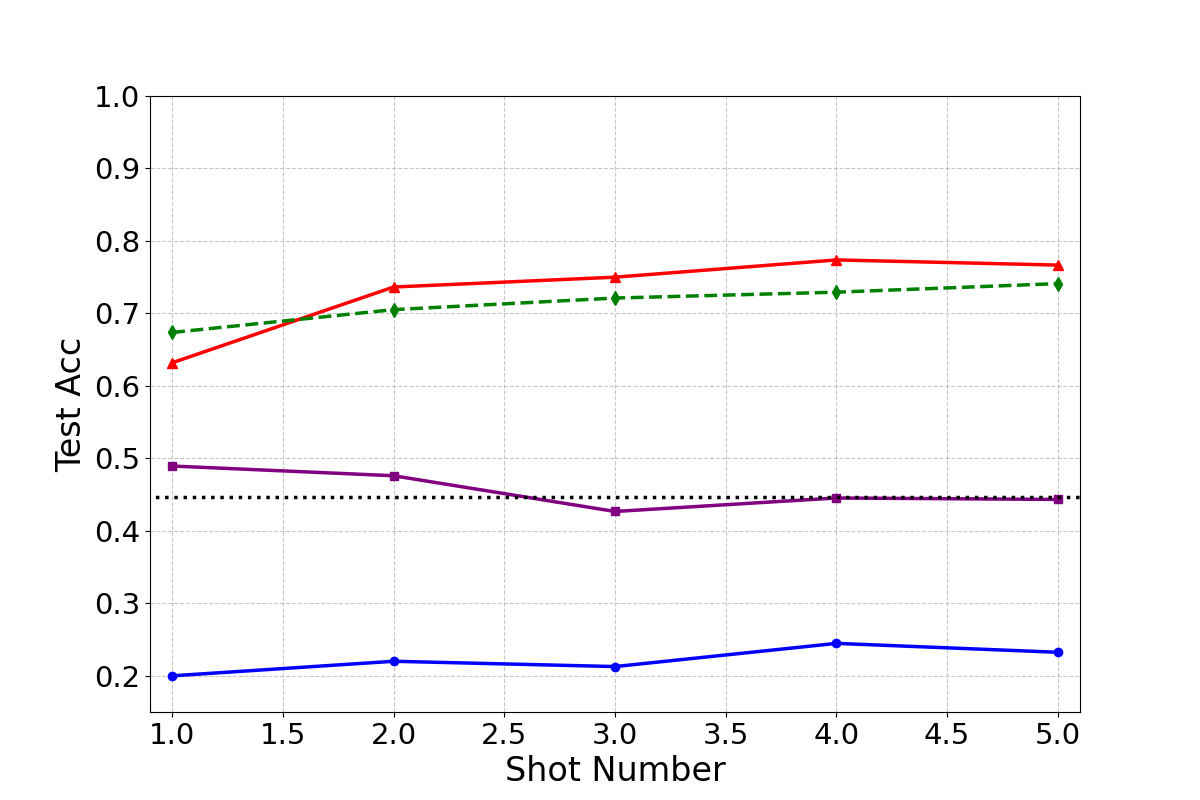}} \\
\subfloat[Cross-Room Scenario]{\includegraphics[width=0.4\textwidth]{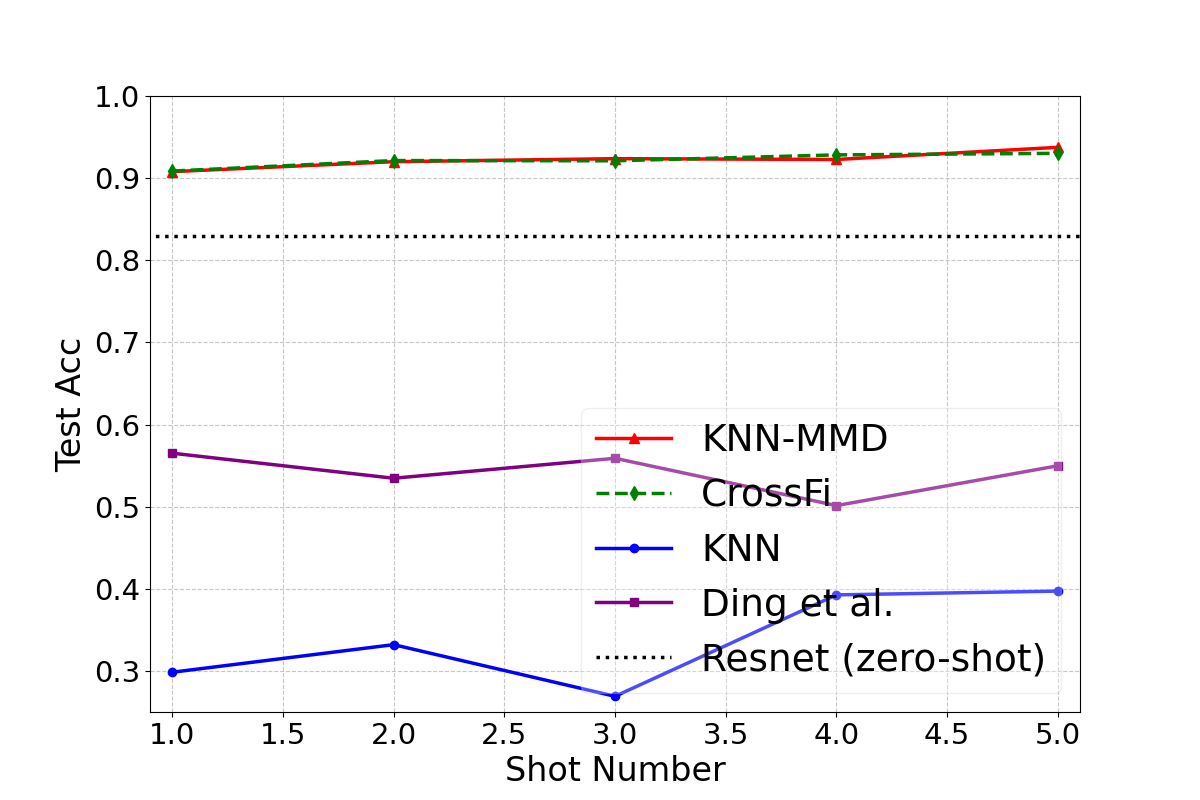}}
\caption{{\color{black} Experiment Results on WiDar3.0 \cite{zhang2021widar3}}}
\label{widar}
\end{figure}

\textcolor{black}{The experimental results are shown in Fig. \ref{widar}, where our KNN-MMD demonstrates excellent performance in both cross-people and cross-room tasks, slightly outperforming CrossFi and significantly outperforming other benchmarks. Since AutoFi \cite{AutoFi} only supports CSI input due to its pre-training method, we have omitted it from this section.}


\subsubsection{Influence of Model Input}

In many Wi-Fi sensing studies, processed CSI signals, such as Time of Flight (ToF), Angle of Arrival (AoA), Attenuation, and \textcolor{black}{Doppler Frequency Shift (DFS)}, are commonly used instead of the original CSI. These processed signals serve as important feature extraction methods in model-based approaches. However, this practice remains controversial in the field of deep learning. While data pre-processing methods can help extract meaningful features, they may also lead to the loss of information present in the original data. As noted in \cite{zhang2021widar3}, DFS contains the most information about velocity distribution, offering advantages over other features. Additionally, Micro-Doppler has been successfully used in OFDM radar sensing tasks and has also shown promising results in Wi-Fi sensing \cite{tang2023mdpose}. 

To investigate this, we compare the performance of KNN-MMD on the WiGesture dataset using both original CSI signals, DFS, and Micro-Doppler. Specifically, DFS is obtained by estimating the frequency shift from the phase differences of consecutive CSI samples, which primarily captures macro motions and change. For a CSI stream denoted as $H(t) = |H(t)| e^{j \angle H(t)}$, the phase difference between consecutive samples is used to estimate the Doppler shift $f_d$ via: $f_d = \frac{1}{2\pi} \frac{\mathrm{d}(\angle H(t))}{\mathrm{d}t}$. \textcolor{black}{Considering the CSI sequence $x \in \mathbb{R}^{l \times s}$, where $l$ is the number of frames and $s$ is the number of subcarriers, we obtain the DFS feature with dimensions $\mathbb{R}^{(l-1) \times s}$. We then expand this feature to include an additional dimension for input into ResNet, resulting in dimensions of $\mathbb{R}^{1 \times (l-1) \times s}$.}

Micro-Doppler, in contrast, is extracted by applying time-frequency analysis such as the Short-Time Fourier Transform (STFT) to the CSI series, which reveals finer-grained motion patterns. \textcolor{black}{Specifically, we set the frequency bins to $f$ to obtain a frequency feature with dimensions $\mathbb{R}^{s \times f \times t}$, where $t$ denotes the time frames after the STFT. We then treat the subcarrier dimension $s$ as the channel and input it into the ResNet for feature extraction. In the experiment, we set the window size to 60, the overlap to 58, and the number of frequency bins to 128.}

\begin{figure}[htbp]
\centering
\includegraphics[width=0.4\textwidth]{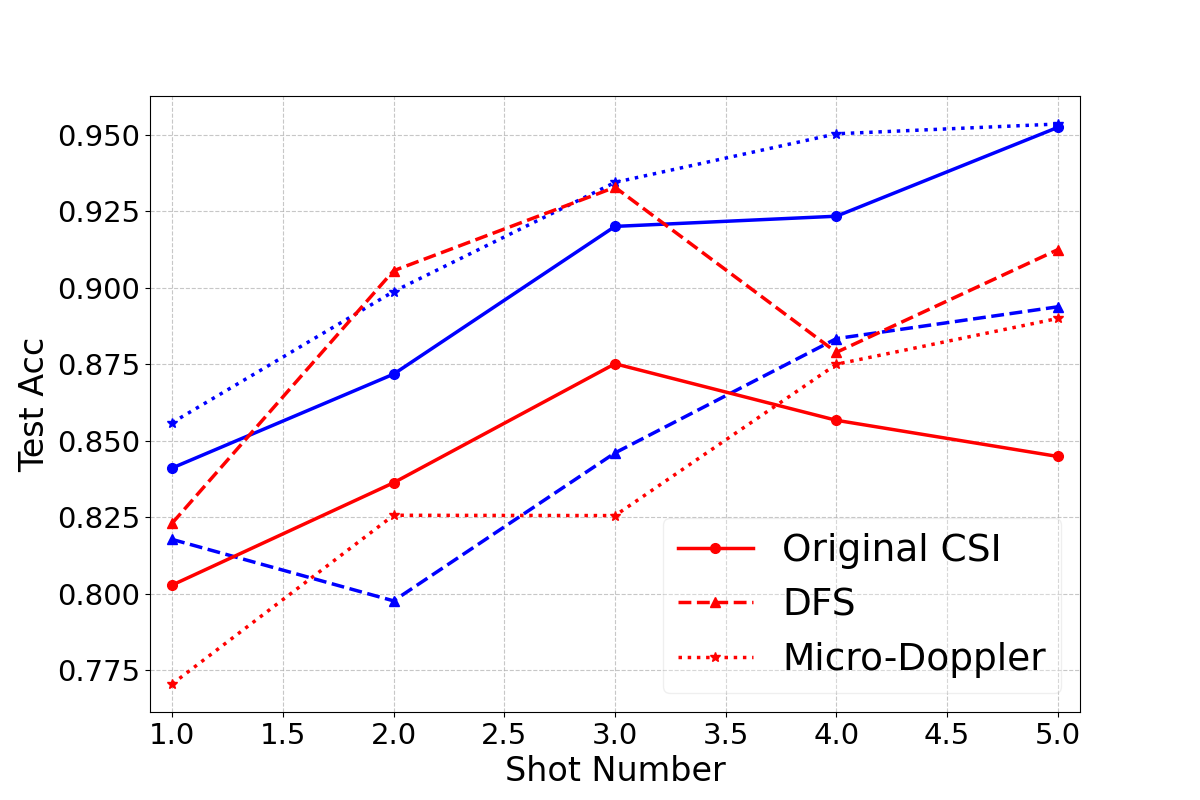}
\caption{{\color{black}Experiment Results on the WiGesture Dataset: The blue lines represent the gesture recognition tasks, while the red lines indicate the people identification tasks.}}
\label{dfs}
\end{figure}

\textcolor{black}{Shown as Fig. \ref{dfs}, the results indicate that, compared to original CSI input, DFS can improve performance in people identification but may lead to lower performance in gesture recognition. Conversely, Micro-Doppler demonstrates an opposite trend. This is because DFS effectively captures the low-frequency, consistent stylistic patterns of an individual's movement, which is stable for identity. In contrast, Micro-Doppler excels at extracting the high-frequency, detailed kinematics of a specific motion, making it superior for classifying gesture types but more sensitive to individual variations that hinder identification.}



\subsection{Sensitivity Analysis}
To evaluate the robustness of our KNN-MMD method, we conduct a series of sensitivity analyses focused on the gesture recognition task, examining three key components: the data dimension of UMAP dimension reduction \(d\), the number of neighbors in KNN \(k\), and the kernel list in MK-MMD \(K\). The results for the data dimension \(d\) and the number of neighbors \(k\) are presented in Table \ref{tab:KNN}. Our findings indicate that these factors have only a minimal impact on model performance.

Regarding the kernel lists, this aspect was mentioned only briefly in previous works. Here, we evaluate two popular types of kernel functions: the Gaussian kernel and the Laplacian kernel, defined as follows:
\begin{equation}
\begin{aligned}
k_G^\sigma (x_1,x_2) &= \exp\left({-\frac{||x_1-x_2||_2^2}{2\sigma^2}} \right), \\
k_L^\sigma (x_1,x_2) &= \exp\left({-\frac{||x_1-x_2||_1}{\sigma}} \right),
\end{aligned}
\end{equation}
where \(k_G^\sigma\) represents the Gaussian kernel with parameter \(\sigma\) based on Gaussian distance, and \(k_L^\sigma\) represents the Laplacian kernel with parameter \(\sigma\) based on Manhattan distance. To construct the kernel function list, we select two kernels of each type, with hyperparameters \(\sigma\) set at 0.5 and 1.0. For the mixture kernel list, we employ one Gaussian kernel and one Laplacian kernel, both with \(\sigma = 1.0\). The experimental results are shown in Table \ref{tab:kernel}, where we observe that the Gaussian kernels used in our experiments demonstrate superior performance in the one-shot gesture recognition task on both the WiGesture and WiDar3.0 datasets. However, these parameters are akin to hyperparameters in neural network training, such as learning rate and batch size, and it is not universally conclusive which settings are the best across all scenarios. They should be adapted appropriately based on the specific problem and task context.

\begin{table}[htbp]
\caption{\color{black}{One-shot Gesture Recognition}}
    \centering
    \begin{tabular}{|c||c|c|}
        \hline
        \textbf{Kernels} & \textbf{WiGesture} \cite{CSI-BERT} & \textbf{WiDar3.0} \cite{zhang2021widar3} \\
        \hline \hline
        \textbf{Gaussian Kernels} & \textbf{93.26\%} & \textbf{66.79\%} \\
        \hline
        \textbf{Laplacian Kernels} & 91.81\% & 54.51\% \\
        \hline
        \textbf{Mixture Kernels} & 93.07\% & 66.02\% \\
        \hline
    \end{tabular}
\label{tab:kernel}
\end{table}

\begin{figure}[htbp]
\centering 
\includegraphics[width=0.4\textwidth]{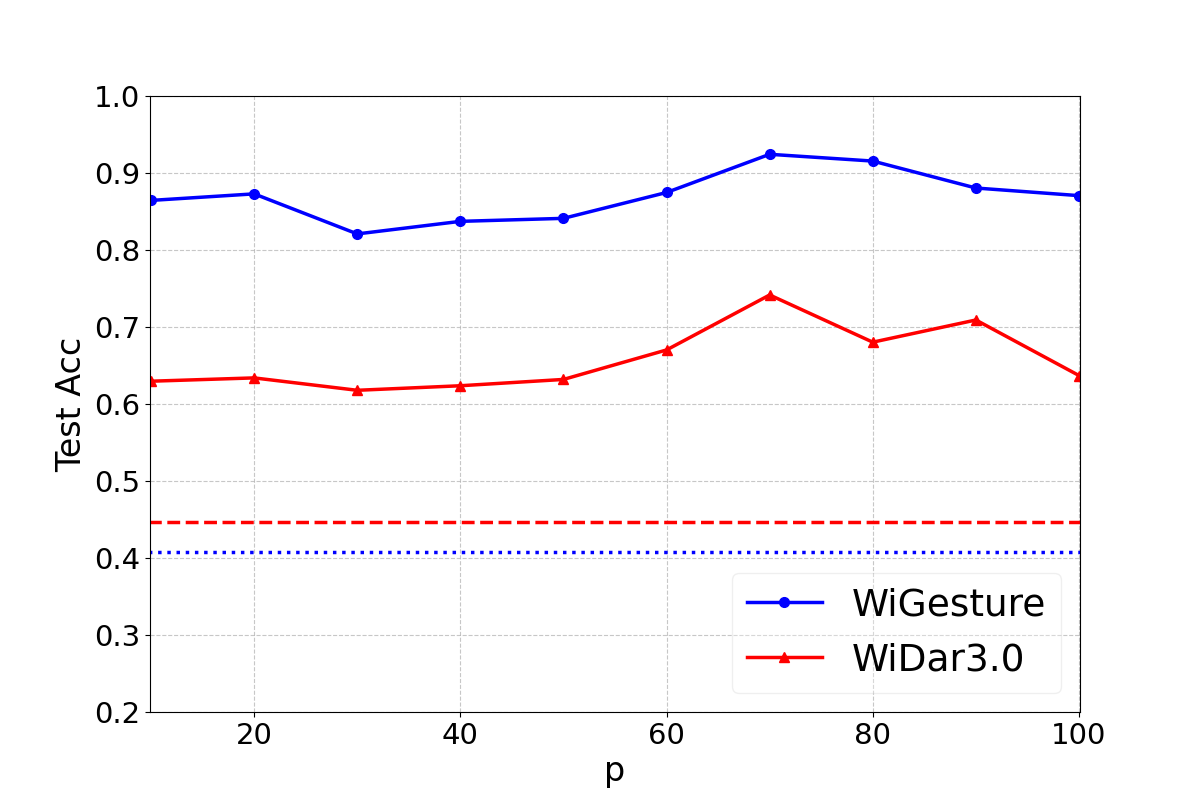}
\caption{{\color{black} One-shot Gesture Recognition: The axis represents the top \(p\%\) samples in the testing set used to construct the help set. The dashed and dotted lines indicate the benchmarks for the WiDar3.0 \cite{zhang2021widar3} and WiGesture \cite{CSI-BERT} datasets, respectively.}}
\label{fig:top}
\end{figure}

Additionally, when integrating them into our KNN-MMD method, the proportion \(p\%\) used to construct the help set from the testing set is also an important hyperparameter. To illustrate its influence on our method, we conduct a series of experiments in the one-shot gesture recognition task on the WiGesture and WiDar3.0 datasets, where \(p\%\) is selected from \(10\%\) to \(100\%\). The experimental results are shown in Fig. \ref{fig:top}. We observe that accuracy first increases and then decreases as \(p\%\) grows. When \(p\%\) is small, the sample size in the help set is also small. However, MK-MMD relies on the statistical properties of both the training and help sets, and when the data amount is too little, it introduces random error. Conversely, when \(p\%\) is too large, the quality of pseudo labels in the help set diminishes, affecting the performance of local alignment. Surprisingly, the accuracy does not decrease sharply even when \(p\%\) is very large. One potential reason is that the alignment achieved by MK-MMD is based on the statistical characteristics of the feature space. Even if there are some incorrect labels in the target domain, the correct labels occupy a larger proportion, limiting the impact of the incorrect labels on the statistical properties. Additionally, the loss function incorporates not only the alignment term but also the classification term, which mitigates the influence of incorrect pseudo labels. Overall, the accuracy shows only a minor variation across different \(p\%\) values, remaining within a \(10\%\) change, which is acceptable in practice.

\color{black}

\section{Conclusion} \label{Conclusion}

In this paper, we introduce KNN-MMD, a FSL method for cross-domain Wi-Fi sensing. We demonstrate that traditional DAL has its shortcomings and propose a local distribution alignment method to address this problem. Additionally, our method supports an early stop strategy to identify when to stop training, which is hardly used in most traditional FSL methods. The experimental results show that our model performs well and remains stable across various cross-domain WiFi sensing tasks, including gesture recognition, people identification, fall detection, and action recognition. The stability of our model also indicates its potential for practical applications.

In the future, there are several aspects of our method that can be further explored. Firstly, our method is not limited to KNN and MMD, and it would be interesting to investigate the effectiveness of other combinations of metric-based methods or few-shot learning methods with DAL methods. Additionally, by combining a zero-shot learning method with the DAL framework, we could also expand its range of applications. Moreover, the current framework could be used only for classification tasks. It would be worth exploring its application in regression tasks.

\bibliographystyle{ieeetr}
\bibliography{ref.bib}

\section*{Biography}
\begin{IEEEbiography}[{\includegraphics[width=1in,height=1.25in,clip,keepaspectratio]{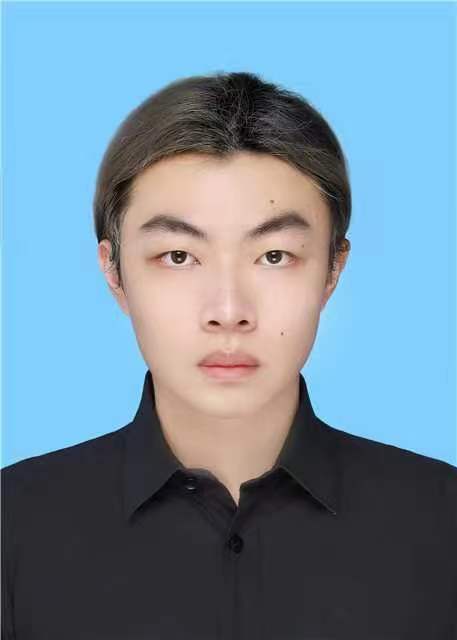}}]{Zijian Zhao} (Graduate Student Member, IEEE)
received the B.Eng. degree in computer science and technology from School of Computer Since and Engineering, Sun Yat-sen University in 2024. He is currently pursuing the Ph.D. degree in civil engineering (scientific computation) with Department of Civil and Environmental Engineering, The Hong Kong University of Science and Technology. He was a visiting student in Shenzhen Research Institute of Big Data, The Chinese University of Hong Kong (Shenzhen) from 2023 to 2024. His current research interests include deep learning, reinforcement learning, intelligent transportation, mobile computing, and wireless sensing.
\end{IEEEbiography}

\begin{IEEEbiography}[{\includegraphics[width=1in,height=1.25in,clip,keepaspectratio]{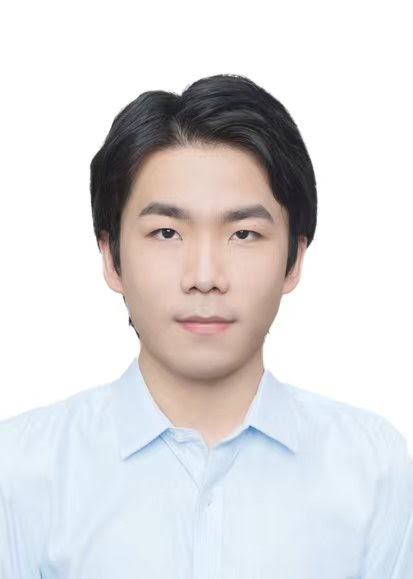}}]{Zhijie Cai} (Graduate Student Member, IEEE)
received the B.Sc. degree in mathematics and applied mathematics from Sun Yat-sen University in 2022. He is currently pursuing the Ph.D. degree with the Shenzhen Research Institute of Big Data, and the School of Science and Engineering, The Chinese University of Hong Kong, Shenzhen. His current research interests include edge intelligence, wireless sensing, and deep learning.
\end{IEEEbiography}

\begin{IEEEbiography}[{\includegraphics[width=1in,height=1.25in,clip,keepaspectratio]{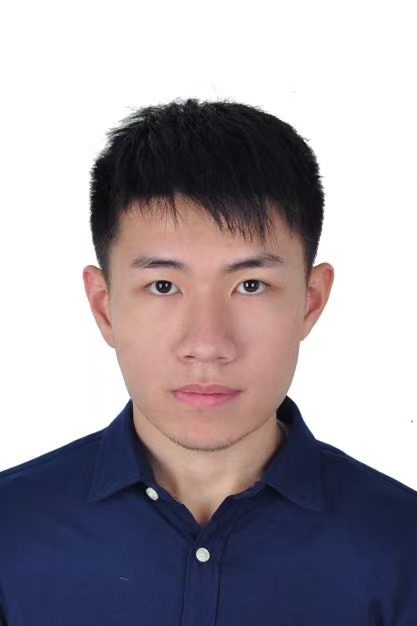}}]{Tingwei Chen}
is currently a visiting student at the Shenzhen Research Institute of Big Data. He received his M.Sc. degree in communication engineering from The Chinese University of Hong Kong, Shenzhen, in 2023 and his B.S. degree in electronic information science and technology from Sun Yat-sen University in 2021. His research interests include multimodal machine learning and wireless sensing.
\end{IEEEbiography}

\begin{IEEEbiography}[{\includegraphics[width=1in,height=1.25in,clip,keepaspectratio]{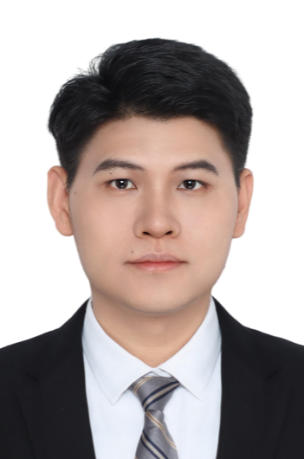}}]{Xiaoyang Li} (Member, IEEE) is currently an Assistant Professor with Southern University of Science and Technology. He received the Ph.D. degree from The University of Hong Kong. His research interests include integrated sensing-communication-computation and edge learning. He is a recipient of Young Elite Scientists Sponsorship Program by CAST, Forbes China 30 under 30, Young Elite of G20, Overseas Youth Talent in Guangdong, Overseas High-caliber Personnel in Shenzhen, Outstanding Research Fellow in Shenzhen, the Best Paper Award of IEEE 4th International Symposium on Joint Communications and Sensing, the Exemplary Reviewers of IEEE Wireless Communications Letters and Journal of Information and Intelligence (JII). He has served as the Editor of JII, and the Workshop Chairs of IEEE ICASSP/WCNC/PIMRC/MIIS/MediCom.
\end{IEEEbiography}

\begin{IEEEbiography}[{\includegraphics[width=1in,height=1.25in,clip,keepaspectratio]{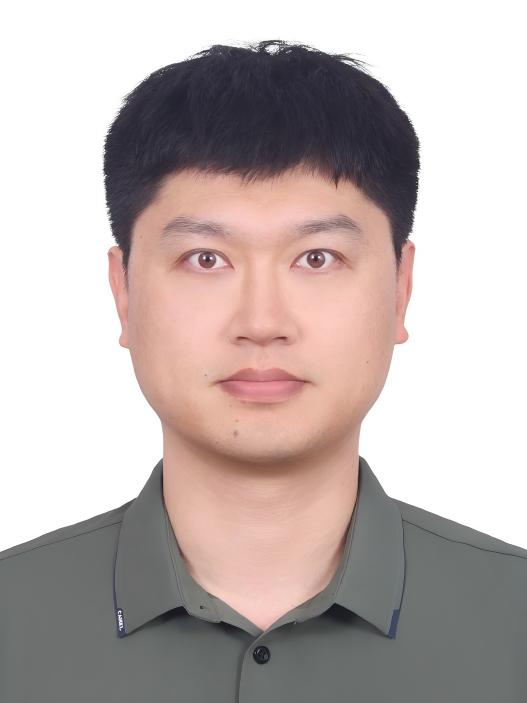}}]{Hang Li}
received the B.E. and M.S. degrees from Beihang University, Beijing, China, in 2008 and 2011, respectively, and the Ph.D. degree from Texas A\&M University, College Station, TX, USA, in 2016. He was a postdoctoral research associate with both Texas A\&M University and University of California-Davis (Sept. 2016-Mar. 2018). He was a visiting research scholar (Apr. 2018 – June 2019) and a research scientist (June 2019 – May 2025) at Shenzhen Research Institute of Big Data, Shenzhen, China. His current research interests include wireless networks, Internet of things, stochastic optimization, and applications of machine learning. He is recognized as Overseas High-Caliber Personnel (Level C) at Shenzhen in 2020. 
\end{IEEEbiography}

\begin{IEEEbiography}[{\includegraphics[width=1in,height=1.25in,clip,keepaspectratio]{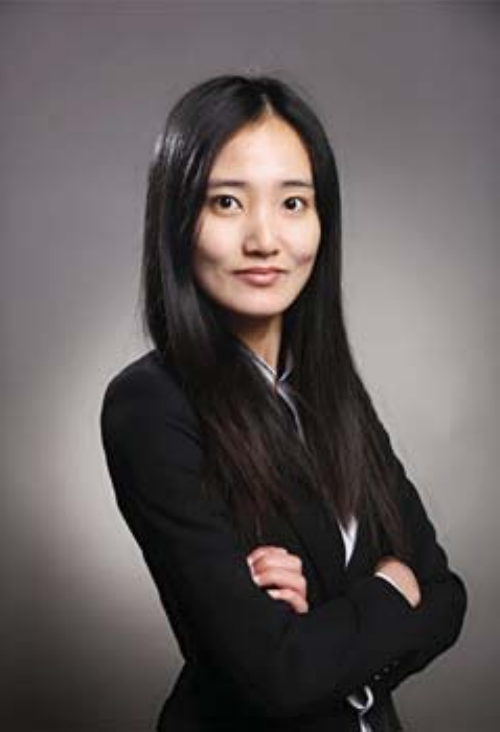}}]{Qimei Chen}
(Member, IEEE) received the Ph.D. degree from the College of Information Science and Electronic Engineering, Zhejiang University, Hangzhou, China, in 2017. She was a Visiting Student with the Department of Electrical and Computer Engineering, University of California at Davis, Davis, CA, USA, from 2015 to 2016. From 2017 to 2022, she was an Associate Researcher with the School of Electric Information, Wuhan University, Wuhan, China, where she has been an Associate Professor, since 2023. Her research interests include intelligent edge communication, massive MIMO, and machine learning in wireless communications. She received the Exemplary Reviewer Certificate of the IEEE Wireless Communications Letters in 2020 and 2023. She has served as a workshop co-chair and TPC Member for IEEE conferences, such as ICC, GLOBECOM, PIMRC, and WCNC.
\end{IEEEbiography}

\begin{IEEEbiography}[{\includegraphics[width=1in,height=1.25in,clip,keepaspectratio]{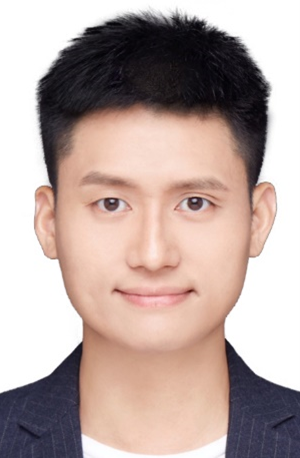}}]{Guangxu Zhu}
(Member, IEEE) received the Ph.D. degree in electrical and electronic engineering from The University of Hong Kong in 2019. Currently he is a senior research scientist and deputy director of network system optimization center at the Shenzhen research institute of big data, and an adjunct associate professor with the Chinese University of Hong Kong, Shenzhen. His recent research interests include edge intelligence, semantic communications, and integrated sensing and communication. He is a recipient of the 2023 IEEE ComSoc Asia-Pacific Best Young Researcher Award and Outstanding Paper Award, the World's Top 2\% Scientists by Stanford University, the "AI 2000 Most Influential Scholar Award Honorable Mention", the Young Scientist Award from UCOM 2023, the Best Paper Award from WCSP 2013 and IEEE  JSnC 2024. He serves as associate editors at top-tier journals in IEEE, including IEEE TMC, TWC and WCL. He is the vice co-chair of the IEEE ComSoc Asia-Pacific Board Young Professionals Committee.
\end{IEEEbiography}


 




\vfill

\end{document}